%% file: main.tex
\documentclass[11pt]{article}

\usepackage[preprint]{acl}

\usepackage{times}
\usepackage{latexsym}
\usepackage[T1]{fontenc}
\usepackage[utf8]{inputenc}
\usepackage{microtype}
\usepackage{inconsolata}
\usepackage{graphicx}
\usepackage{booktabs}
\usepackage{amsmath,amssymb,amsfonts,amsthm}
\usepackage{bm}
\usepackage{multirow}
\usepackage{xcolor}
\usepackage{colortbl}
\usepackage{array}
\usepackage{tabularx}
\usepackage{enumitem}
\usepackage[ruled,vlined]{algorithm2e}
\SetKwInput{KwInput}{Input}
\SetKwInput{KwOutput}{Output}

\newtheorem{proposition}{Proposition}
\newtheorem{theorem}{Theorem}

\newtheorem{corollary}{Corollary}

\newtheorem{remark}{Remark}

\definecolor{oursbg}{RGB}{253,238,240}
\definecolor{simred}{RGB}{165,30,45}

\newcommand{\method}{\textsc{Loom}}
\newcommand{\E}{\mathbb{E}}
\newcommand{\R}{\mathbb{R}}
\newcommand{\ip}[2]{\langle #1,#2\rangle_{M}}
\newcommand{\nrm}[1]{\|#1\|_{M}}
\newcommand{\gbar}{\bar g}
\newcommand{\cin}{c^{\mathrm{in}}}
\newcommand{\SNR}{\mathrm{SNR}}
\newcommand{\AM}{\mathrm{AM}}
\newcommand{\HM}{\mathrm{HM}}

\input{tables/stats.tex}

\title{Every Batch Is Its Own Validation Set: Leave-One-Out Gradient\\Matching for Online Data Selection in LLM Fine-Tuning}

\author{
  \textbf{Hongyu Chen\textsuperscript{1}},
  \textbf{Xinyi Luo\textsuperscript{1}},
  \textbf{Ming Zhao\textsuperscript{2}},
  \textbf{Lin Tang\textsuperscript{1}},
  \textbf{Zihan Xu\textsuperscript{2}},
  \\
  \textbf{Jing Li\textsuperscript{1}},
  \textbf{Yuxuan Wang\textsuperscript{2}},
  \textbf{Haoran Deng\textsuperscript{1}},
  \textbf{Wei Zhang\textsuperscript{1}}
  \\
  \\
  \textsuperscript{1}Sichuan University, Chengdu, China \\
  \textsuperscript{2}University of Electronic Science and Technology of China, Chengdu, China
}

\begin{document}
\maketitle

\begin{abstract}
Online batch selection fine-tunes a language model on the most useful part of each candidate batch. Selectors that match the gradient of the candidate batch are attractive because they need no held-out data, yet they rarely beat training on the whole batch. We show why. In-sample gradient matching uses every example as part of its own target, so its objective credits each example with its own gradient noise. This is the covariance penalty that makes training error optimistic, now sitting on the diagonal of the gradient Gram matrix: it steers selection toward the noisiest examples and makes the full batch the best solution the objective can reach. The fix costs nothing. For each example, the other candidates form an independent sample of the data distribution, so removing the diagonal turns the matching objective into an unbiased estimate of the update's error with respect to the population gradient. The minimizer of this leave-one-out objective weights examples by their gradient signal-to-noise ratio (SNR), and whenever per-example SNR is heterogeneous enough, half of a batch yields a lower-error update than the whole batch; we give the exact condition. We build \method{} on this principle. It computes the Gram matrix in the metric of the Adam preconditioner during the ordinary backward pass, selects a weighted subset greedily with a $(1-e^{-\gamma})$ guarantee, and uses no held-out data. Across four fine-tuning tasks and seven backbones from 1.5B to 8B parameters, \method{} improves on full-batch training by \gainRegL{} and \gainRegQ{} points on Llama-3.1-8B and Qwen2.5-7B, exceeds every in-sample gradient matcher by \gainMatchL{} points and the validation-guided GREATS and OPUS by \gainOPUSL{}--\gainGREATSQ{}, and selects injected label noise at under a fifth of its base rate.
\end{abstract}

\section{Introduction}
\label{sec:intro}

Supervised fine-tuning (SFT) adapts a pretrained language model with a few epochs over task data~\citep{hu2022lora,zhou2023lima,ivison2023camels}, and the quality of what the model sees at each step matters as much as how much it sees~\citep{albalak2024survey,zhang2025survey,zou2025utility}. \emph{Online batch selection} acts on this at training time: every step draws a candidate batch and updates the model on a subset of it~\citep{loshchilov2015online,katharopoulos2018not}. Existing selectors either score examples by loss or gradient norm~\citep{jiang2019accelerating}, compare against a reference model~\citep{mindermann2022prioritized,evans2024data,wang2026blade}, align candidates with the gradient of a held-out validation or proxy set~\citep{wang2024greats,wang2026opus,wang2026filter}, or rely on validation-free utility and diversity signals~\citep{hong2024diversified,zou2025utility}.

Gradient matching is the most principled of the validation-free options. It chooses a weighted subset whose gradient reproduces the gradient of the candidate batch, typically through facility location or orthogonal matching pursuit with approximation guarantees~\citep{mirzasoleiman2020coresets,killamsetty2021gradmatch,nguyen2025minibatch}. Its goal also contains a puzzle. If the subset reproduced the batch gradient exactly, the update would equal the full-batch update, which is already available once the batch has been back-propagated. The objective thus treats the whole batch as its ceiling, and in our experiments every in-sample matcher lands between random selection and full-batch training (\S\ref{sec:main}).

\begin{figure*}[t]
\centering
\includegraphics[width=\textwidth]{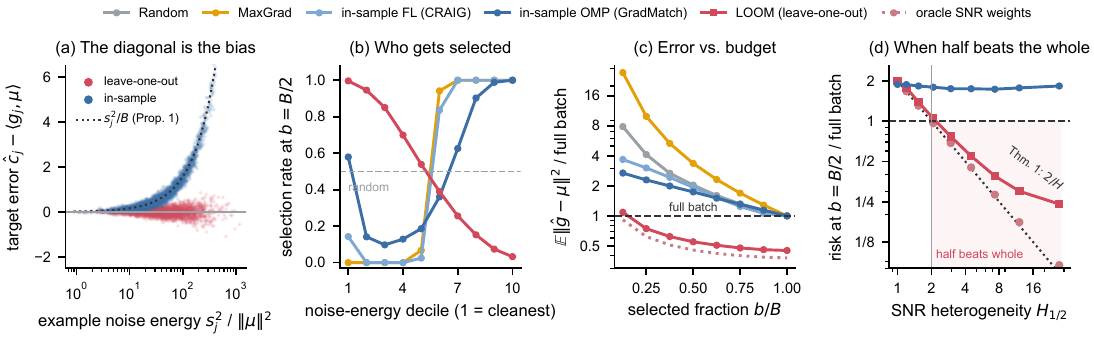}
\caption{\textbf{The diagonal is the bias} (exact computation on the signal-plus-noise model of \S\ref{sec:optimum}; $B{=}64$, log-normal noise). (a) In-sample targets over-estimate an example's alignment with the population gradient by its noise energy divided by $B$ (Prop.~\ref{prop:optimism}); leave-one-out targets are unbiased. (b) At $b{=}B/2$, in-sample matchers and MaxGrad select the noisiest examples, while \method{} selects the cleanest. (c) Error of the selected update relative to the full batch: in-sample matching never beats the full batch, whereas leave-one-out matching does at every $b/B\ge 1/4$. (d) At $b{=}B/2$ the best achievable risk ratio is $2/H_{1/2}$ (Thm.~\ref{thm:half}), so half of a batch beats the whole once the SNR heterogeneity $H_{1/2}$ exceeds 2.}
\label{fig:motivation}
\end{figure*}

We argue that the ceiling comes from the target. In-sample matching uses the batch both as data and as target, so each example appears in the gradient it is matched to. In the Gram matrix $K_{ij}=\langle g_i,g_j\rangle$ of per-example gradients, the in-sample target of example $j$ is $\frac1B\sum_iK_{ij}$, which contains its self-similarity $K_{jj}/B$. We show that this term is exactly the covariance penalty that makes training error an optimistic estimate of test error~\citep{efron2004estimation,stein1981estimation}. It credits each example with its own gradient noise, so in-sample matching prefers the noisiest examples (Figure~\ref{fig:motivation}b) and its optimum reproduces the sampling noise of the batch instead of estimating the population gradient.

The fix is to use every candidate batch as its own validation set. For each example, the other $B-1$ candidates form an independent sample of the data distribution, so the leave-one-out mean $\frac{1}{B-1}\sum_{i\neq j}K_{ij}$ estimates the example's alignment with the population gradient without bias. Dropping the diagonal thus turns the matching objective into an unbiased estimate of the update's error, with three consequences. The optimal weights are proportional to each example's gradient signal-to-noise ratio (SNR), shrunk as the batch's signal weakens. Unlike in-sample matching, the objective can beat the whole batch, exactly when a heterogeneity index of per-example SNR exceeds $B/b$ (Figure~\ref{fig:motivation}d). And the error it estimates is the only selector-dependent term in the convergence bound of preconditioned SGD.

We turn this principle into \method{} (\emph{Leave-One-Out Matching}). \method{} computes per-example LoRA gradients in the metric of the Adam preconditioner during the ordinary backward pass, forms leave-one-out targets, selects a weighted subset greedily with a $(1-e^{-\gamma})$ guarantee, and updates on the weighted subset; it adds 5\% to the step time and needs no validation data. Our contributions:
\begin{itemize}[leftmargin=*,itemsep=1pt,topsep=2pt]
\item \textbf{Diagnosis.} We show that in-sample gradient matching is optimistic by a covariance penalty on the Gram diagonal, which makes it select noisy examples and caps it at the full batch (\S\ref{sec:theory}).
\item \textbf{Principle.} Leave-one-out targets give an unbiased estimate of the population risk. We derive its optimum, the exact condition under which a subset beats the whole batch, a greedy guarantee, and a convergence bound in which the estimated risk is the error term (\S\ref{sec:theory}--\ref{sec:method}).
\item \textbf{Evidence.} On four SFT tasks and seven backbones, \method{} improves on full-batch training by \gainRegL{}--\gainRegQ{} points and on all in-sample matchers by \gainMatchL{}, is competitive with or better than validation-guided selectors, and filters injected label noise. Mechanism measurements on real batches match the theory (\S\ref{sec:experiments}).
\end{itemize}

\section{Related Work}
\label{sec:related}

\paragraph{Online batch selection.}
Early selectors prioritize high-loss examples or sample by gradient norm~\citep{loshchilov2015online,jiang2019accelerating,katharopoulos2018not}. RHO-LOSS scores points by the gap to a reference model trained on held-out data~\citep{mindermann2022prioritized}; JEST and ACID select jointly with a reference model~\citep{evans2024data,udandarao2025active}, and BLADE keeps that reference synchronized with training~\citep{wang2026blade}. For LLMs, GREATS maximizes a Taylor approximation of the validation-loss decrease with ghost inner products~\citep{wang2024greats}, OPUS projects optimizer-shaped updates onto a proxy direction~\citep{wang2026opus}, and Filter-then-Weight matches a target-oriented update under the optimizer state~\citep{wang2026filter}. Validation-free selectors include DivBS, which maximizes orthogonalized representativeness~\citep{hong2024diversified}, and UDS, which combines the nuclear norm of the logits with embedding diversity~\citep{zou2025utility}. \method{} is validation-free and gradient-based, and differs from these methods in the target it matches.

\paragraph{Gradient matching and coresets.}
CRAIG selects medoids of the gradient geometry by facility location~\citep{mirzasoleiman2020coresets}, GRAD-MATCH matches the training gradient by orthogonal matching pursuit~\citep{killamsetty2021gradmatch}, CREST builds coresets from mini-batches~\citep{yang2023towards}, CRUST uses medoids for robustness to noisy labels~\citep{mirzasoleiman2020crust}, and CoLM adapts mini-batch coresets to LLMs with Adam-normalized gradients~\citep{nguyen2025minibatch}. Offline variants cluster or pursue gradients of instruction data~\citep{zhang2024tagcos,deng2024influential,chang2026spice}. All of them target the in-sample batch or dataset gradient, which is the target Proposition~\ref{prop:optimism} identifies as optimistic. Their guarantees rest on (weak) submodularity~\citep{nemhauser1978analysis,das2011submodular,elenberg2018restricted}, which our objective retains.

\paragraph{Self-influence, coherence, and gradient noise.}
Per-example gradients are affordable through the machinery developed for differential privacy~\citep{goodfellow2015efficient,li2022large,bu2023differentially} and in-run data attribution~\citep{wang2025data,ding2026inrun}. A large self-influence $\|g_j\|^2$ flags mislabeled or outlying examples~\citep{pruthi2020estimating,thakkar2023self,chhabra2025outlier}, generalization is attributed to directions shared across examples~\citep{chatterjee2020coherent,zielinski2020weak}, and gradient agreement filtering drops conflicting micro-batches~\citep{chaubard2024beyond}. Stochastic gradient noise is heavy-tailed in deep networks, attention models, and LLM training~\citep{simsekli2019tail,zhang2020why,yu2026open}. \method{} connects these observations. The self-influence that flags noise is exactly the diagonal that in-sample matching rewards, and removing it leaves the cross-example agreement that the coherence view identifies as signal. Offline selection for instruction tuning~\citep{xia2024less,liu2024what,li2024quantity,yang2024smalltolarge} is reviewed in Appendix~\ref{app:related}.

\section{Gradient Matching Overfits the Batch}
\label{sec:theory}

\paragraph{Setup.}
At a training step with parameters $\theta$, a candidate batch $z_1,\dots,z_B$ is drawn i.i.d.\ from the training distribution. Let $g_j=\nabla\ell(\theta;z_j)$ be the per-example gradients and $\mu=\E[g_j]$ the population gradient. The optimizer applies a metric $M\succ0$ that is fixed before the batch is drawn: $M=I$ for SGD, and $M=\mathrm{diag}(\sqrt{\hat v}+\epsilon)^{-1}$ for Adam-type methods~\citep{kingma2015adam} with the second-moment estimate $\hat v$ of the previous step (\S\ref{sec:method}). We write $\ip{u}{v}=u^\top Mv$ and $K_{ij}=\ip{g_i}{g_j}$. A selector returns weights $w\ge0$ supported on $S\subseteq[B]$ with $|S|\le b$, and the update direction is $\hat g(w)=\sum_j w_jg_j$. Its error is
\begin{equation}
\begin{aligned}
R(w)&=\nrm{\hat g(w)-\mu}^2\\
&=w^\top Kw-2\textstyle\sum_j w_j\ip{g_j}{\mu}+\nrm{\mu}^2 .
\end{aligned}
\label{eq:risk}
\end{equation}
Theorem~\ref{thm:conv} shows that $R$ is the only selector-dependent term in the convergence bound. In Eq.~\eqref{eq:risk}, only the alignments $\ip{g_j}{\mu}$ are unobserved, so a matching selector is defined by how it estimates them. In-sample matching~\citep{mirzasoleiman2020coresets,killamsetty2021gradmatch,hong2024diversified,nguyen2025minibatch} replaces $\mu$ by the batch mean $\gbar$ and minimizes $\nrm{\hat g(w)-\gbar}^2=w^\top Kw-2w^\top\cin+\mathrm{const}$ with $\cin_j=\frac1B\sum_iK_{ij}$. Leave-one-out matching uses $c_j=\frac{1}{B-1}\sum_{i\neq j}K_{ij}$ instead:
\begin{equation}
\hat R(w)=w^\top Kw-2w^\top c .
\label{eq:loo}
\end{equation}

\paragraph{The optimism of in-sample matching.}
The two targets differ only through the diagonal of $K$, and that difference is a systematic bias. Proofs are in Appendix~\ref{app:proofs}.

\begin{proposition}[The diagonal is the optimism]
\label{prop:optimism}
(i) $\E[c_j\mid z_j]=\ip{g_j}{\mu}$. (ii) $\cin_j=c_j+\frac1B(K_{jj}-c_j)$, so $\E[\cin_j-\ip{g_j}{\mu}\mid z_j]=\frac1B\ip{g_j}{g_j-\mu}$. (iii) For every weighting $w_j=\omega(z_j)$ that depends on each example alone, $\E\hat R(w)=\E R(w)-\nrm{\mu}^2$, whereas the in-sample objective satisfies $\E[\nrm{\hat g(w)-\gbar}^2]=\E R(w)-\nrm{\mu}^2+\E\nrm{\gbar}^2-\frac2B\E\sum_j\omega(z_j)\ip{g_j}{g_j-\mu}$.
\end{proposition}

The last term in (iii) is the covariance penalty of \citet{efron2004estimation}, the gap between the training error and the test error of a fitted model. On average $\E\ip{g}{g-\mu}=\mathrm{tr}(M\Sigma)$, the noise energy of a single gradient, so in-sample matching underestimates the error of any selector in proportion to the weight it places on noisy examples. Two failure modes follow. First, the objective prefers noisy examples, which is exactly the self-influence signal that flags mislabeled data~\citep{pruthi2020estimating}; on synthetic batches it selects the noisiest decile every time (Figure~\ref{fig:motivation}b). Second, its minimizer over all weights is $w=\frac1B\bm 1$, which reproduces $\gbar$ together with its noise, so in-sample matching can at best recover the full batch. Facility location has the same flaw, because every selected point covers itself with its maximal similarity (Appendix~\ref{app:fl}).

\paragraph{What the unbiased objective prefers.}
\label{sec:optimum}
To see what $\hat R$ rewards, consider heteroscedastic signal-plus-noise gradients $g_j=a_j\mu+\xi_j$, where each example has a signal loading $a_j$ and a noise $\xi_j$ with mean zero, $\E\nrm{\xi_j}^2=s_j^2$, independent across examples. Its per-example SNR is $\SNR_j=a_j^2m/s_j^2$ with $m=\nrm{\mu}^2$. Given the attributes, $\E[K\mid a,s]$ has entries $a_ia_jm$ off the diagonal and $a_j^2m+s_j^2$ on it, and the risk is $\E[R(w)\mid a,s]=m\big(\sum_jw_ja_j-1\big)^2+\sum_jw_j^2s_j^2$, which $\hat R$ estimates without bias up to a constant (Proposition~\ref{prop:optimism}).

\begin{proposition}[Signal-to-noise weights with shrinkage]
\label{prop:optimum}
Under this model, the minimizer of $\E[R(w)\mid a,s]$ over $w$ supported on $S$ is
\begin{equation}
w^\star_j=\frac{m}{1+mQ_S}\,\frac{a_j}{s_j^2},\qquad Q_S=\textstyle\sum_{i\in S}a_i^2/s_i^2 ,
\label{eq:wstar}
\end{equation}
with risk $m/(1+\SNR_S)$ and $\SNR_S=\sum_{j\in S}\SNR_j$. The attained reduction $G(S)=m\,\SNR_S/(1+\SNR_S)$ is monotone submodular and is maximized by the $b$ examples with the largest $\SNR_j$.
\end{proposition}

The optimum is inverse-variance weighting along the signal direction, $w_j\propto a_j/s_j^2$. Its total signal $\sum_jw^\star_ja_j=\SNR_S/(1+\SNR_S)<1$ shrinks the step when the batch carries little signal, as a James--Stein estimator would~\citep{stein1981estimation}. The next result states when a subset of this form beats the whole batch.

\begin{theorem}[When half a batch beats the whole]
\label{thm:half}
Let $a_j\equiv1$ and normalize the weights to $\sum_jw_j=1$. The best $b$-subset has risk $m/(b\,\AM_b)$, where $\AM_b$ is the arithmetic mean of the $b$ largest $\SNR_j$, and the uniform full batch has risk $m/(B\,\HM_B)$, where $\HM_B$ is the harmonic mean of all $\SNR_j$. Selection therefore strictly improves on the full batch if and only if
\begin{equation}
H_\rho:=\AM_b/\HM_B>1/\rho,\qquad \rho=b/B .
\label{eq:hetero}
\end{equation}
The in-sample objective, in contrast, is minimized by the uniform full batch.
\end{theorem}

Because $H_\rho\ge1$ with equality only for identical noise levels, selection helps only when per-example noise is heterogeneous. Heavy-tailed gradient noise, typical of attention models and LLM training~\citep{zhang2020why,yu2026open}, supplies this heterogeneity: on real SFT batches $H_{1/2}$ lies between 2 and 6 (\S\ref{sec:mechanism}), consistent with reports that a well-chosen subset can outperform full-data fine-tuning~\citep{zou2025utility}. Monte Carlo risks match Theorem~\ref{thm:half} within 8\% (Figure~\ref{fig:motivation}d, Table~\ref{tab:numerics}).

\paragraph{The risk controls convergence.}
The quantity that $\hat R$ estimates is also the error term in the convergence bound of the optimizer.

\begin{theorem}[Selected updates]
\label{thm:conv}
Let $f$ be $L$-smooth and bounded below by $f^\star$, and let $\theta_{t+1}=\theta_t-\eta M_t\hat g_t$ with metrics satisfying $M_t\preceq\epsilon^{-1}I$. For $\eta\le\epsilon/(4L)$ and any selector,
\begin{equation}
\begin{aligned}
\frac1T\sum_{t<T}\E\|\nabla f(\theta_t)\|_{M_t}^2&\le\frac{4(f(\theta_0)-f^\star)}{\eta T}\\
&\quad+\frac3T\sum_{t<T}\E R_t ,
\end{aligned}
\label{eq:conv}
\end{equation}
where $R_t=\|\hat g_t-\nabla f(\theta_t)\|_{M_t}^2$.
\end{theorem}

The selector enters only through its average risk measured in the optimizer's metric. The full batch has $\E R_t=\mathrm{tr}(M_t\Sigma_t)/B$, in-sample matching approaches this value as its matching error vanishes, and leave-one-out matching minimizes an estimate of $R_t$ that is unbiased up to a constant.

\section{\method{}}
\label{sec:method}

\begin{figure*}[t]
\centering
\includegraphics[width=\textwidth]{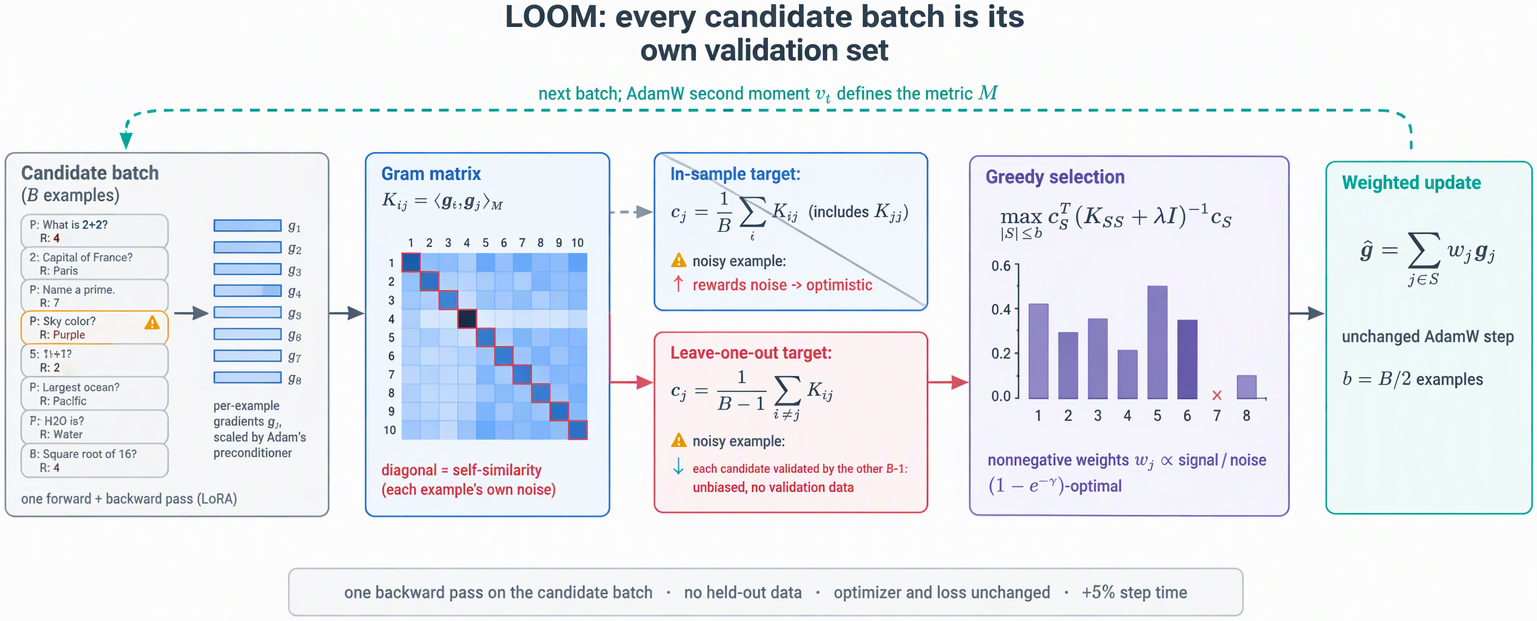}
\caption{\textbf{\method{} in one training step.} One backward pass yields per-example LoRA gradients, scaled by the Adam preconditioner, and their Gram matrix $K$. In-sample targets count its diagonal as signal; \method{} validates each candidate against the other $B{-}1$, selects a weighted subset greedily, and takes an unchanged AdamW step.}
\label{fig:overview}
\end{figure*}

\method{} replaces the choice of which examples enter the update and leaves the loss and optimizer unchanged (Figure~\ref{fig:overview}, Algorithm~\ref{alg:loom}).

\paragraph{Gram matrix in the optimizer's metric.}
With LoRA, per-example gradients are small ($r(d_{\mathrm{in}}{+}d_{\mathrm{out}})$ entries per module) and are formed during the ordinary backward pass, as in the LoRA implementation of GREATS~\citep{wang2024greats}. We scale them by $M_t^{1/2}=(\sqrt{\hat v_{t-1}}+\epsilon)^{-1/2}$, using AdamW's second moment from the previous step so that the metric is fixed before the batch is drawn, and accumulate $K=\tilde G\tilde G^\top$ module by module. Measuring geometry in the optimizer's metric follows recent optimizer-aware selectors~\citep{wang2026opus,wang2026filter,ding2026inrun}; our contribution concerns the target, not the metric. For dense modules with short sequences, ghost inner products compute the same $K$ without materializing gradients~\citep{li2022large,wang2024greats} (Appendix~\ref{app:impl}).

\paragraph{Objective and greedy selection.}
With leave-one-out targets $c$ and a ridge $\lambda=\bar\lambda\,\mathrm{tr}(K)/B$, a support $S$ reduces the estimated risk by
\begin{equation}
\begin{aligned}
G(S)&=\max_{w}\big\{2w^\top c_S-w^\top(K_{SS}+\lambda I)w\big\}\\
&=c_S^\top(K_{SS}+\lambda I)^{-1}c_S .
\end{aligned}
\label{eq:G}
\end{equation}
The ridge keeps the weights from fitting noisy targets. We maximize $G$ by forward selection. Each step adds the candidate with the largest marginal gain among those whose residual correlation $c_j-K_{jS}w_S$ is positive, and the weights on $S$ solve the nonnegative version of Eq.~\eqref{eq:G}.

\begin{proposition}[Greedy guarantee]
\label{prop:greedy}
$G$ is monotone, and its submodularity ratio satisfies $\gamma\ge\lambda_{\min}^{(2b)}(K{+}\lambda I)/\lambda_{\max}^{(b+1)}(K{+}\lambda I)\ge\lambda/(\lambda+\lambda_{\max}(K))$, where $\lambda^{(k)}$ denote extreme eigenvalues over $k\times k$ principal submatrices. Forward selection therefore returns $S$ with $G(S)\ge(1-e^{-\gamma})\max_{|S'|\le b}G(S')$.
\end{proposition}

The bound follows from restricted strong concavity~\citep{das2011submodular,elenberg2018restricted}. It is conservative in practice: on exhaustively searchable batches, greedy selection attains on average 99.99\% of the optimum and never less than 99.6\%, for both signed and nonnegative weights (Table~\ref{tab:numerics}).

\begin{table*}[t]
\centering
\footnotesize
\setlength{\tabcolsep}{4.1pt}
\renewcommand{\arraystretch}{0.97}
\input{tables/main_results.tex}
\caption{\textbf{Main results} (accuracy or pass@1, \%, mean of four seeds; standard deviations in Table~\ref{tab:seeds}). Every selector keeps $b{=}64$ of $B{=}128$ candidates per step; Regular updates on all 128. \textbf{Ext.}: external resource (ref.: reference model; val.: validation set; proxy: held-out proxy set). Groups: full batch; validation-free scoring; validation-free in-sample gradient matching; external resources; ours. \textbf{Bold}/\underline{underline}: best/second best.}
\label{tab:main}
\end{table*}

\paragraph{Update.}
The selected weights are normalized to $\sum_jw_j=1$, so that AdamW sees a weighted mean with the scale of an ordinary step; Prop.~\ref{prop:optimum} predicts that the step would additionally shrink as the batch SNR falls, which we found unnecessary under Adam (Table~\ref{tab:ablation-full}). Because the per-example gradients are already stored, the update $\sum_{j\in S}w_jg_j$ needs no second backward pass. With gradient accumulation, the per-example gradients of all micro-batches are kept until the step and selection runs once per optimizer step.

\begin{algorithm}[t]
\footnotesize
\DontPrintSemicolon
\SetAlgoLined
\caption{\method{}: one optimizer step}
\label{alg:loom}
\KwInput{candidate batch $\{z_j\}_{j=1}^B$; budget $b$; ridge $\bar\lambda$; AdamW state $\hat v_{t-1}$}
Forward and backward pass on the batch; keep per-example LoRA gradients $g_j$\;
$\tilde g_j\leftarrow(\sqrt{\hat v_{t-1}}+\epsilon)^{-1/2}\odot g_j$;\quad $K\leftarrow\tilde G\tilde G^\top$\;
$c_j\leftarrow\frac{1}{B-1}\sum_{i\neq j}K_{ij}$;\quad $\lambda\leftarrow\bar\lambda\,\mathrm{tr}(K)/B$\;
$S\leftarrow\emptyset$\;
\While{$|S|<b$ \textbf{and} $\exists\, j\notin S$ with $r_j=c_j-K_{jS}w_S>0$}{
  $j^\star\leftarrow\arg\max_{j:\,r_j>0}\ r_j^2/(K_{jj}+\lambda-k_{jS}^\top(K_{SS}+\lambda I)^{-1}k_{jS})$\;
  $S\leftarrow S\cup\{j^\star\}$;\quad $w_S\leftarrow\arg\min_{w\ge0}\ w^\top(K_{SS}+\lambda I)w-2w^\top c_S$\;
}
AdamW step on $\hat g=\sum_{j\in S}w_jg_j/\sum_{j\in S}w_j$\;
\end{algorithm}

\paragraph{Cost and special cases.}
For Llama-3.1-8B with rank-8 LoRA on all linear modules, per-example gradients have 21M entries, about 0.7\,GB per GPU for a 128-example step on eight GPUs; forming $K$ costs $B^2$ inner products, and the greedy loop costs $O(Bb^2)$. The measured overhead is \ovhLoom\% of the step time (Table~\ref{tab:cost}). Two existing selectors are special cases of the same quadratic objective. Keeping the diagonal in the target gives in-sample matching, and replacing $c$ by the alignment with a validation gradient while fixing equal weights gives the first-order GREATS objective with an identity Hessian~\citep{wang2024greats} (Appendix~\ref{app:greats}). \method{} is the self-validated member of this family.

\section{Experiments}
\label{sec:experiments}

\subsection{Setup}
\label{sec:setup}

\begin{figure*}[t]
\centering
\includegraphics[width=\textwidth]{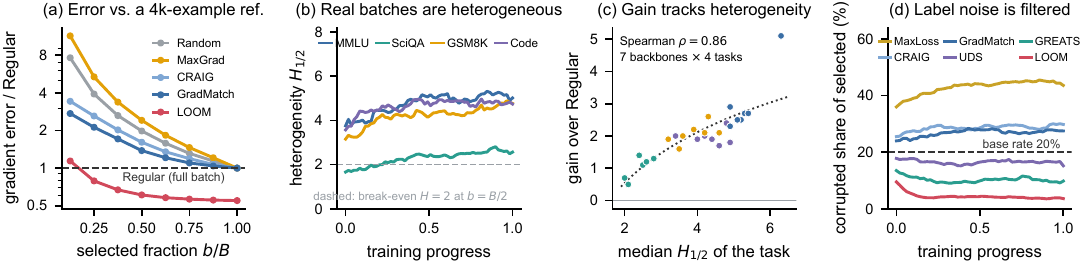}
\caption{\textbf{Mechanism on real batches} (Llama-3.1-8B unless noted). (a) Error of the selected update against a reference gradient from 4{,}096 held-out training examples (MMLU, step 400), relative to Regular. (b) SNR heterogeneity $H_{1/2}$ of candidate batches during training. (c) Gain of \method{} over Regular against the task's median $H_{1/2}$ for seven backbones and four tasks (colors as in b). (d) Share of corrupted examples among the selected ones when 20\% of responses are corrupted.}
\label{fig:mechanism}
\end{figure*}

\paragraph{Tasks, models, and training.}
We follow the SFT protocol of prior online-selection work~\citep{zou2025flylora,zou2025utility}. Models are trained on the auxiliary training set of MMLU and evaluated on its test set~\citep{hendrycks2021measuring}, trained and evaluated on ScienceQA~\citep{lu2022learn} and GSM8K~\citep{cobbe2021training}, and trained on CodeAlpaca-20k~\citep{chaudhary2023code} and evaluated on HumanEval (pass@1)~\citep{chen2021evaluating}. All evaluations are zero-shot, and we report the mean over four seeds. The main backbones are Llama-3.1-8B~\citep{grattafiori2024llama} and Qwen2.5-7B~\citep{yang2024qwen25}; Qwen2.5-1.5B/3B, Llama-3.2-3B, Mistral-7B-v0.3~\citep{jiang2023mistral}, and Qwen3-8B-Base~\citep{yang2025qwen3} test generality. We train rank-8 LoRA adapters ($\alpha{=}16$) on all linear projections~\citep{hu2022lora} with AdamW~\citep{loshchilov2019decoupled}, using the per-task epochs, learning rates, and sequence lengths of \citet{zou2025utility}. Each optimizer step draws $B{=}128$ candidates in micro-batches of 8, and every selector keeps $b{=}64$ ($\rho{=}1/2$).

\paragraph{Baselines.}
\emph{Regular} updates on the whole candidate batch and \emph{Random} on a uniform half. Validation-free scoring baselines are MaxLoss~\citep{loshchilov2015online}, MaxGrad~\citep{katharopoulos2018not}, and UDS~\citep{zou2025utility}. The in-sample gradient matchers are CRAIG~\citep{mirzasoleiman2020coresets}, GradMatch~\citep{killamsetty2021gradmatch}, DivBS~\citep{hong2024diversified}, and CoLM~\citep{nguyen2025minibatch}. Methods with external resources are RHO-Loss with a larger reference model~\citep{mindermann2022prioritized}, GREATS with five validation examples per task~\citep{wang2024greats}, and OPUS with a proxy set of 256 held-out training examples~\citep{wang2026opus}. All gradient-based methods use \method{}'s per-example gradients in the Adam metric, so they differ from \method{} only in objective and target. \method{} uses $\bar\lambda{=}0.1$ on every task and backbone. Appendix~\ref{app:setup} lists all settings.

\subsection{Main Results}
\label{sec:main}

\method{} attains the best average on both backbones (Table~\ref{tab:main}): \avgLoomL{} on Llama-3.1-8B and \avgLoomQ{} on Qwen2.5-7B, which is $+\gainRegL$ and $+\gainRegQ$ over Regular. Three patterns stand out. First, every in-sample matcher lands between Random and Regular (\matchMinL--\matchMaxL{} and \matchMinQ--\matchMaxQ{}), as Proposition~\ref{prop:optimism} predicts: the more exactly they match the batch, the closer they get to Regular, and none exceeds it. GradMatch uses the same gradients and metric and a closely related greedy solver, yet \method{} is $\gainGradMatchL$ and $\gainGradMatchQ$ points better. Second, \method{} exceeds the validation-guided GREATS and OPUS by \gainOPUSL--\gainGREATSQ{} points without held-out data. In single-task SFT, the training distribution is also the test distribution, and each leave-one-out target averages 127 in-distribution gradients, whereas GREATS averages five validation gradients. Third, the gains are largest on MMLU, whose auxiliary training set pools several sources and has the most heterogeneous batches (\S\ref{sec:mechanism}), and smallest on ScienceQA, whose examples are short and uniform. UDS is marginally ahead on Qwen2.5-7B ScienceQA and on Llama-3.1-8B GSM8K. On five further backbones from 1.5B to 8B parameters, \method{} improves on Regular by \extraGainMin--\extraGainMax{} points and on UDS by \extraUDSMin--\extraUDSMax{} (Table~\ref{tab:extra}).

\begin{table}[t]
\centering
\footnotesize
\setlength{\tabcolsep}{2.6pt}
\input{tables/ablation.tex}
\caption{\textbf{Ablations} on Llama-3.1-8B (four-task average). \textbf{Err.}: error of the selected update against a 4{,}096-example reference gradient, relative to Regular (lower is better). Further variants are in Table~\ref{tab:ablation-full}.}
\label{tab:ablation}
\end{table}

\subsection{Ablations}
\label{sec:ablation}

Table~\ref{tab:ablation} changes one component at a time. \emph{The target is what matters.} Keeping the diagonal, all else unchanged, loses \ablInSample{} points and more than doubles the gradient error (1.38 vs.\ 0.61), nearly the whole gap to GradMatch. The momentum $m_{t-1}$ is out-of-sample but stale and recovers only part of the gap; two-fold cross-fitting is unbiased but averages half as many gradients per target ($-0.32$). A Euclidean metric costs 0.82, because a few modules dominate its geometry (Appendix~\ref{app:extra}). Redundancy-aware selection beats ranking by SNR, in line with the value of diversity in batch selection~\citep{hong2024diversified,zou2025utility}. Signed weights lower the gradient error (0.58) yet lose 1.51 points, because negative weights take ascent steps on single examples, which a first-order risk does not penalize. Reweighting the whole batch matches selecting half of it, since selection is the sparse limit of optimal weighting (Proposition~\ref{prop:optimum}), and the forward-only \method{}-fast keeps 72\% of the gain.

\subsection{Mechanism Analysis}
\label{sec:mechanism}

\begin{figure*}[t]
\centering
\includegraphics[width=\textwidth]{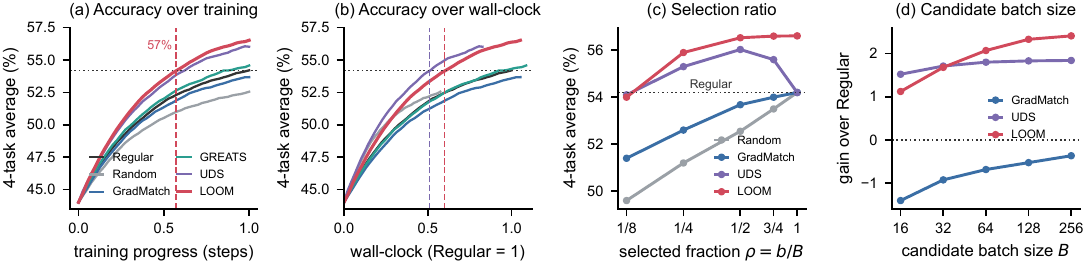}
\caption{\textbf{Training dynamics and budgets} (Llama-3.1-8B, four-task average). (a) Accuracy over training; dashed: \method{} reaches Regular's final accuracy (dotted) after 57\% of the steps. (b) The same runs over wall-clock time. (c) Selected fraction $\rho$ ($\rho{=}1$ is Regular for all but \method{}, which reweights the batch). (d) Candidate batch size $B$ at $\rho{=}1/2$: gain over Regular trained on the same batches.}
\label{fig:results}
\end{figure*}

\paragraph{Leave-one-out beats the batch on real gradients.}
Against a reference gradient from 4{,}096 held-out training examples (Figure~\ref{fig:mechanism}a), in-sample matchers approach Regular from above as $b$ grows, as in the synthetic Figure~\ref{fig:motivation}c, whereas \method{} stays below Regular for every $b/B\ge1/4$ (0.61 at $b{=}B/2$).

\paragraph{Real batches are heterogeneous, and the gain tracks it.}
We estimate per-example SNR from $K$ alone (Appendix~\ref{app:snr}). The index $H_{1/2}$ lies between 2 and 6 and grows during training as the mean gradient shrinks relative to per-example noise (Figure~\ref{fig:mechanism}b); ScienceQA, with short and uniform examples, stays near the break-even value of 2. Across seven backbones and four tasks, the gain of \method{} over Regular increases with the task's median $H_{1/2}$ (Spearman $\rho=\rhoHgain$; Figure~\ref{fig:mechanism}c), as Theorem~\ref{thm:half} predicts.

\paragraph{Noisy examples are filtered.}
When 20\% of the responses are replaced by responses to other prompts, MaxLoss, CRAIG, and GradMatch select corrupted examples above the base rate (\corrMaxLoss\%, \corrCRAIG\%, \corrGM\%), as Proposition~\ref{prop:optimism} implies. GREATS and UDS select them below it (\corrGREATS\%, \corrUDS\%), and \method{} selects \corrLoom\% (Figure~\ref{fig:mechanism}d): a corrupted response has a large diagonal entry and a small leave-one-out target. At 40\% corruption \method{} loses \noiseDropLoom{} points, against \noiseDropReg{} for Regular and \noiseDropGM{} for GradMatch, and it leads GREATS, whose validation examples are clean, by 3.2 points (Table~\ref{tab:noise}).

\subsection{Efficiency and Generality}
\label{sec:generality}

\paragraph{Efficiency.}
\method{} runs at \thrLoom$\times$ the throughput of Regular; its \ovhLoom\% overhead goes to per-example gradients, the Gram matrix, and the greedy loop (Tables~\ref{tab:cost} and~\ref{tab:stepms}). It reaches Regular's final accuracy after \stepsLoom\% of the steps and \timeLoom\% of the wall-clock time (Figure~\ref{fig:results}a,b). UDS back-propagates only the selected half and is therefore faster (\thrUDS$\times$ throughput, \timeUDS\% of the wall-clock), but it ends \gainUDSL{} points lower. \method{}-fast also back-propagates only the selected half and reaches \thrFast$\times$ throughput with an average of \avgFastL.

\paragraph{Budgets.}
\method{} beats Regular at every $\rho\ge1/4$ and saturates at $\rho{=}1/2$, whereas UDS peaks at $\rho{=}1/2$ and in-sample matching rises monotonically toward Regular (Figure~\ref{fig:results}c). As $B$ grows from 16 to 256, GradMatch's deficit shrinks from $-1.41$ to $-0.37$, the $1/B$ decay of the optimism in Proposition~\ref{prop:optimism}, while \method{}'s gain grows from $+1.12$ to $+2.41$ because its targets average more gradients (Figure~\ref{fig:results}d).

\paragraph{Beyond single-task SFT.}
Appendix~\ref{app:extra} covers five further settings. In a GREATS-style protocol with a mixed instruction pool and small target-task validation sets~\citep{xia2024less,wang2024greats}, GREATS wins on TydiQA~\citep{clark2020tydi}, whose validation set carries information the pool lacks, while \method{} leads on MMLU and BBH~\citep{suzgun2023challenging}; the two are within 0.2 points on average, and adding the validation gradient to \method{}'s target is best overall. When the four training sets are pooled, the smallest source (GSM8K, 5\%) is under-weighted, as CoLM observed for in-sample matching~\citep{nguyen2025minibatch}, which per-source targets fix. The gains also hold for full fine-tuning, for transfer from GSM8K to MATH500~\citep{hendrycks2021math,lightman2024verify} and SVAMP~\citep{patel2021nlp}, and at the per-task selection ratios of \citet{zou2025utility}.

\section{Conclusion}
\label{sec:conclusion}

Validation-free gradient matching evaluates each candidate against a target that contains the candidate itself. The resulting optimism sits on the diagonal of the gradient Gram matrix, and it explains why such selectors favor noisy examples and cannot beat the full batch. Removing the diagonal turns the candidate batch into its own validation set: the objective becomes an unbiased estimate of the update's error, its optimum weights examples by signal-to-noise ratio, and a subset beats the whole batch whenever per-example SNR is heterogeneous, which is the case in LLM fine-tuning. \method{} implements this with one backward pass and a 5\% overhead, and it improves on full-batch training, in-sample matchers, and validation-guided selectors. The same correction applies to any selector whose target is estimated from the data it selects, including offline coresets and data attribution.

\section*{Limitations}

\method{} estimates the gradient of the \emph{training} distribution. When the target task differs from the training pool, a validation gradient carries information the batch does not, and validation-guided selection can be better; Appendix~\ref{app:extra} shows such a case and a simple combination. In pooled data with rare sources, leave-one-out targets favor the majority unless they are computed per source. Our analysis assumes i.i.d.\ candidates, a metric that is fixed within a step, and finite noise variance; with duplicated examples, a candidate's near-copies act as part of its own target, and deduplication or grouped leave-one-out is needed. Per-example gradients are cheap for LoRA, but full fine-tuning of large models requires ghost inner products or sketches. Because SNR weights favor low-noise examples, they correlate mildly with response length, and online selection can shift model behavior in ways that task accuracy does not reveal~\citep{zeng2026online}. Our experiments cover SFT of models up to 8B parameters; pretraining and reinforcement-learning post-training remain to be tested.

\section*{Ethics Statement}

\method{} changes only which training examples enter each update, so it inherits the risks of the models and data it is applied to. Its tendency to down-weight examples that disagree with the rest of a batch filters noise, but it could also down-weight rare but legitimate content, such as minority dialects or uncommon viewpoints; practitioners should audit selected data for such effects, and the stratified variant in Appendix~\ref{app:extra} limits the effect for known sources. All datasets and models are public and used under their licenses, and no human subjects or personal data are involved. AI assistants were used for language editing and for checking LaTeX; the method, analysis, and experimental design are the authors' own, and every reference was checked against its primary source.

\bibliography{custom}

\appendix
\input{appendix.tex}

\end{document}

%% file: tables/stats.tex
\newcommand{\avgLoomL}{56.5}
\newcommand{\avgLoomQ}{70.9}

\newcommand{\gainRegL}{2.3}
\newcommand{\gainRegQ}{2.4}
\newcommand{\gainUDSL}{0.5}

\newcommand{\gainGREATSQ}{2.0}
\newcommand{\gainOPUSL}{1.6}

\newcommand{\matchMinL}{53.4}
\newcommand{\matchMaxL}{54.2}
\newcommand{\matchMinQ}{67.7}
\newcommand{\matchMaxQ}{68.5}
\newcommand{\gainMatchL}{2.4}

\newcommand{\gainGradMatchL}{2.9}
\newcommand{\gainGradMatchQ}{2.8}
\newcommand{\extraGainMin}{1.7}
\newcommand{\extraGainMax}{1.9}
\newcommand{\extraUDSMin}{0.7}
\newcommand{\extraUDSMax}{0.8}
\newcommand{\rhoHgain}{0.86}
\newcommand{\noiseDropLoom}{3.2}
\newcommand{\noiseDropReg}{6.9}
\newcommand{\noiseDropGM}{7.4}

\newcommand{\corrLoom}{3.6}
\newcommand{\corrGM}{27.5}
\newcommand{\corrCRAIG}{29.8}
\newcommand{\corrMaxLoss}{43.6}
\newcommand{\corrGREATS}{9.8}
\newcommand{\corrUDS}{15.2}
\newcommand{\ovhLoom}{5.4}
\newcommand{\thrLoom}{0.95}
\newcommand{\stepsLoom}{57}
\newcommand{\timeLoom}{60}

\newcommand{\timeUDS}{51}
\newcommand{\thrFast}{1.17}
\newcommand{\thrUDS}{1.21}
\newcommand{\avgFastL}{55.9}

\newcommand{\ablInSample}{2.79}

%% file: tables/main_results.tex
\begin{tabular}{l c ccccc ccccc}
\toprule
 & & \multicolumn{5}{c}{\textbf{Llama-3.1-8B}} & \multicolumn{5}{c}{\textbf{Qwen2.5-7B}} \\
\cmidrule(lr){3-7}\cmidrule(lr){8-12}
\textbf{Method} & \textbf{Ext.} & MMLU & SciQA & GSM8K & HumanE. & Avg. & MMLU & SciQA & GSM8K & HumanE. & Avg. \\
\midrule
Regular & -- & 38.3 & 93.2 & 56.0 & 29.3 & 54.2 & 55.3 & 94.6 & 78.2 & 45.8 & 68.5 \\
\midrule
Random & -- & 35.6 & 92.9 & 54.9 & 26.8 & 52.6 & 54.6 & 93.5 & 77.8 & 41.3 & 66.8 \\
MaxLoss & -- & 35.7 & 92.8 & 55.4 & 27.2 & 52.8 & 54.8 & 93.2 & 77.9 & 42.1 & 67.0 \\
MaxGrad & -- & 35.9 & 92.8 & 55.1 & 26.9 & 52.7 & 54.7 & 93.9 & 77.7 & 41.6 & 67.0 \\
UDS & -- & \underline{40.1} & \underline{94.3} & \textbf{58.9} & \underline{30.8} & \underline{56.0} & \underline{59.6} & \textbf{95.2} & \underline{79.6} & \underline{46.2} & \underline{70.2} \\
\midrule
CRAIG & -- & 37.2 & 93.0 & 55.4 & 28.0 & 53.4 & 55.0 & 94.1 & 77.9 & 43.9 & 67.7 \\
GradMatch & -- & 37.6 & 93.1 & 55.7 & 28.3 & 53.7 & 55.4 & 94.3 & 78.1 & 44.5 & 68.1 \\
DivBS & -- & 37.9 & 93.3 & 55.9 & 28.5 & 53.9 & 55.7 & 94.4 & 78.2 & 44.8 & 68.3 \\
CoLM & -- & 38.2 & 93.4 & 56.2 & 28.8 & 54.2 & 56.0 & 94.5 & 78.4 & 44.9 & 68.5 \\
\midrule
RHO-Loss & ref. & 37.6 & 93.4 & 56.5 & 27.2 & 53.7 & 56.6 & 93.8 & 78.4 & 43.3 & 68.0 \\
GREATS & val. & 39.0 & 93.7 & 57.0 & 28.6 & 54.6 & 57.5 & 94.2 & 78.7 & 45.1 & 68.9 \\
OPUS & proxy & 39.3 & 93.8 & 57.3 & 29.2 & 54.9 & 57.9 & 94.4 & 79.0 & 45.6 & 69.2 \\
\midrule
\rowcolor{oursbg}\method{} & -- & \textbf{41.2} & \textbf{94.6} & \underline{58.6} & \textbf{31.7} & \textbf{56.5} & \textbf{60.4} & \underline{95.1} & \textbf{80.4} & \textbf{47.6} & \textbf{70.9} \\
\bottomrule
\end{tabular}

%% file: tables/ablation.tex
\begin{tabular}{l c c c}
\toprule
\textbf{Variant} & \textbf{Avg.} & $\bm{\Delta}$ & \textbf{Err.} \\
\midrule
\rowcolor{oursbg}\method{} (full) & \textbf{56.53} & --- & 0.61 \\
\midrule
\multicolumn{4}{l}{\textit{Target}} \\
\quad in-sample target (keeps diagonal) & 53.74 & $-2.79$ & 1.38 \\
\quad momentum target $m_{t-1}$ & 55.34 & $-1.19$ & 0.86 \\
\quad 2-fold cross-fitted target & 56.21 & $-0.32$ & 0.68 \\
\midrule
\multicolumn{4}{l}{\textit{Metric}} \\
\quad Euclidean metric ($M{=}I$) & 55.71 & $-0.82$ & 0.74 \\
\midrule
\multicolumn{4}{l}{\textit{Selection and weights}} \\
\quad top-$b$ by SNR (no redundancy) & 55.92 & $-0.61$ & 0.69 \\
\quad uniform weights on $S$ & 55.81 & $-0.72$ & 0.72 \\
\quad signed weights & 55.02 & $-1.51$ & 0.58 \\
\midrule
\multicolumn{4}{l}{\textit{Budget and kernel}} \\
\quad reweight all ($b{=}B$) & 56.61 & $+0.08$ & 0.55 \\
\quad \method{}-fast (LM-head sketch) & 55.88 & $-0.65$ & 0.71 \\
\bottomrule
\end{tabular}

%% file: appendix.tex

\section{Extended Related Work}
\label{app:related}

\paragraph{Online batch selection and training-time curation.}
Online batch selection was introduced as rank-based sampling of high-loss examples~\citep{loshchilov2015online}. Selective backpropagation skips the backward pass for low-loss examples~\citep{jiang2019accelerating}, and importance sampling with an upper bound on the gradient norm reduces the variance of an unbiased estimator of the batch gradient~\citep{katharopoulos2018not}. These criteria reward large losses or gradients, which also flags corrupted data. RHO-LOSS therefore subtracts the loss of a reference model trained on held-out data~\citep{mindermann2022prioritized}; Bayesian data selection replaces the reference model by a posterior over a lightweight proxy~\citep{deng2023towards}; JEST and ACID select batches jointly with a pretrained reference for multimodal pretraining~\citep{evans2024data,udandarao2025active}; and BLADE turns the bilevel objective behind influence-based selection into an excess-loss criterion with a reference model that is synchronized with training~\citep{wang2026blade}. Rho-1 applies the excess-loss idea to tokens~\citep{lin2024rho}, and learnability-based batch selection has also been adapted to machine-translation fine-tuning~\citep{ghanizadeh2025dynamic}. Validation-guided selectors score candidates by their effect on a held-out objective: GREATS uses a second-order Taylor expansion of the validation loss and ghost inner products~\citep{wang2024greats}, OPUS projects optimizer-shaped updates onto a proxy direction with CountSketch and Boltzmann sampling~\citep{wang2026opus}, Filter-then-Weight casts selection as matching a target-oriented update under the optimizer state~\citep{wang2026filter}, and GAIA models utility globally with Gaussian processes~\citep{wang2026gaia}. Validation-free methods avoid held-out data, which may be unavailable in practice~\citep{peng2024energy}: DivBS maximizes orthogonalized representativeness~\citep{hong2024diversified}, and UDS scores utility and intra-sample diversity by the nuclear norm of the logits and inter-sample diversity against a buffer of recently selected examples, without extra back-propagation~\citep{zou2025utility}. Domain-level online mixing~\citep{jiang2025adaptive} and meta-learned data valuation~\citep{calian2025datarater,du2026difficulty} operate at coarser granularity or with meta-gradients through a held-out objective. Online selection has also been used to target learnable difficulty in reinforcement-learning post-training~\citep{sun2025improving,cui2026learning} and learning progress in vision-language instruction tuning~\citep{chandhok2026learning}. \method{} is validation-free, per-step, and gradient-based. It differs from all of the above in the target it matches: it estimates the population gradient from the batch itself, without the self-term.

\paragraph{Gradient coresets.}
CRAIG bounds the error of a weighted subset by a facility-location objective over gradient distances~\citep{mirzasoleiman2020coresets}. GRAD-MATCH solves the matching problem directly with orthogonal matching pursuit, either for the training gradient or for a validation gradient~\citep{killamsetty2021gradmatch}, and GLISTER solves a bilevel problem with a validation objective~\citep{killamsetty2021glister}. CREST extracts coresets from random mini-batches with a quadratic model of the loss~\citep{yang2023towards}. CRUST shows that medoids of class-wise gradient clusters are robust to noisy labels~\citep{mirzasoleiman2020crust}, and CoLM matches the gradient of larger mini-batches with Adam-normalized, sparsified gradients and keeps all examples of small sources~\citep{nguyen2025minibatch}. For instruction data, TAGCOS clusters gradients and runs OMP per cluster~\citep{zhang2024tagcos}, gradient trajectory pursuit matches gradients along training~\citep{deng2024influential}, SPICE penalizes gradient conflicts in a log-determinant objective~\citep{chang2026spice}, and GRACE updates coresets dynamically with a $k$-NN graph~\citep{tang2026grace}. These methods target an in-sample gradient, of the batch or of the dataset, and Proposition~\ref{prop:optimism} applies to all of them; Appendix~\ref{app:fl} gives the leave-one-out version of facility location. The guarantees rest on submodularity~\citep{nemhauser1978analysis} or weak submodularity~\citep{das2011submodular,elenberg2018restricted}.

\paragraph{Offline data selection for instruction tuning.}
Small curated sets can suffice for alignment~\citep{zhou2023lima}, and quality filters~\citep{chen2024alpagasus}, complexity and diversity scores~\citep{liu2024what}, and instruction-following difficulty~\citep{li2024quantity} select them automatically. Gradient-based methods select data for a target task with low-rank gradient features and Adam-aware influence~\citep{xia2024less}, cluster training trajectories of small models~\citep{yang2024smalltolarge}, use in-context utility with submodular selection~\citep{agarwal2025delift}, distil second-order influence~\citep{nikdan2025efficient}, align training gradients with a task subspace under the coupled geometry of LoRA~\citep{min2026gist}, attribute with on-policy validation targets~\citep{wang2026drift}, solve a weighted independent set on a similarity graph~\citep{zhang2026seed}, or score by the gradient SNR of a small proxy ensemble~\citep{yuan2026uncertainty}. Related signals include influence consensus across tasks~\citep{wu2024icons}, agreement of training dynamics with an anchor set~\citep{zhao2025clues}, and forward-only token saliency~\citep{nagaraj2025trim}. Other work combines loss dynamics with bandits~\citep{lin2025lead}, prunes by weight-norm dynamics~\citep{hu2025donod}, denoises responses~\citep{luo2024robustft}, or selects pretraining data by predictive compression~\citep{shum2025predictive}. Random selection is a strong baseline at scale~\citep{xia2024rethinking}, and gradient similarity has been shown to track answer format more than task semantics~\citep{kim2026form}. Surveys cover this literature~\citep{albalak2024survey,zhang2025survey}. Unlike these methods, \method{} is online: like other online selectors~\citep{wang2024greats,zou2025utility}, it adapts to the current model and needs no separate scoring pass. It also obtains per-example SNR weights as the optimum of an unbiased risk rather than from a proxy ensemble.

\paragraph{Data attribution and per-example gradients.}
Influence functions~\citep{koh2017understanding}, TracIn~\citep{pruthi2020estimating}, TRAK~\citep{park2023trak}, and their LLM-scale variants~\citep{choe2024your,grosse2023studying} attribute predictions to training data. In-run attribution computes Shapley-style values during a single training run with ghost dot products~\citep{wang2025data}, extends them to Adam~\citep{ding2026inrun}, and captures the trajectory dependence of influence with a trajectory-specific leave-one-out~\citep{wang2025capturing}; related work attributes online RL updates to recent buffer records~\citep{hu2025snapshot} and estimates task-level leave-one-out effects with kernel surrogates~\citep{zhang2026efficient}. That leave-one-out removes a point from \emph{training}. Ours removes a point from its own \emph{target} within one batch, a cross-fitting device that leaves training unchanged. The per-example gradient machinery comes from differentially private training~\citep{goodfellow2015efficient,li2022large,bu2023differentially}, whose LLM-scale implementations keep improving~\citep{wang2025flashdp,ullah2026efficient,dadsetan2025reexamining}.

\paragraph{Gradient noise, coherence, and robust aggregation.}
The gradient noise scale relates the useful batch size to the ratio of noise to signal~\citep{mccandlish2018empirical}. Gradient noise in attention models is heavy-tailed~\citep{zhang2020why,simsekli2019tail,yu2026sign}, which is the regime in which per-example SNR is heterogeneous, and whether AdamW itself copes with such noise is an open question~\citep{yu2026open}. The coherent-gradients view attributes generalization to directions shared by many examples and memorization to idiosyncratic ones~\citep{chatterjee2020coherent,zielinski2020weak}. Gradient agreement filtering drops micro-batches whose gradients conflict~\citep{chaubard2024beyond}, gradient clipping can mitigate label noise~\citep{menon2020can}, self-influence reweights pretraining data~\citep{thakkar2023self}, outlier gradients identify detrimental samples~\citep{chhabra2025outlier}, and the spectra of per-sample gradients flag backdoored fine-tuning data~\citep{zhao2026gradsentry}. \method{}'s weights $a_j/s_j^2$ are the estimator-theoretic version of these heuristics: they are the risk-optimal combination of coherent signal and idiosyncratic noise, estimated from the batch without a threshold.

\paragraph{Unbiased risk estimation and cross-fitting.}
The optimism of training error and its covariance-penalty correction are classical~\citep{efron2004estimation,stein1981estimation}, as is the removal of own-observation bias by sample splitting and cross-fitting~\citep{chernozhukov2018double}. The leave-one-out mean of $K$ is a U-statistic~\citep{hoeffding1948class}. To our knowledge, these tools have not been applied to the target of gradient-matching data selection.

\section{Proofs}
\label{app:proofs}

Throughout, the metric $M$ is fixed before the batch is drawn, $z_1,\dots,z_B$ are i.i.d., $g_j=\nabla\ell(\theta;z_j)$, $\mu=\E g_j$, $\Sigma=\mathrm{Cov}(g_j)$, and $\gbar=\frac1B\sum_jg_j$.

\subsection{Proof of Proposition~\ref{prop:optimism}}
\label{app:proof-optimism}

\emph{(i)} $c_j=\frac{1}{B-1}\sum_{i\neq j}\ip{g_j}{g_i}$. Conditional on $z_j$, each $z_i$ with $i\neq j$ is an independent draw from the data distribution, so $\E[\ip{g_j}{g_i}\mid z_j]=\ip{g_j}{\E g_i}=\ip{g_j}{\mu}$, and averaging over $i$ preserves this.

\emph{(ii)} $\cin_j=\frac1B\big(K_{jj}+\sum_{i\neq j}K_{ij}\big)=\frac1B K_{jj}+\frac{B-1}{B}c_j=c_j+\frac1B(K_{jj}-c_j)$. Taking the conditional expectation and using (i), $\E[\cin_j\mid z_j]-\ip{g_j}{\mu}=\frac1B\big(\nrm{g_j}^2-\ip{g_j}{\mu}\big)=\frac1B\ip{g_j}{g_j-\mu}$.

\emph{(iii)} Let $w_j=\omega(z_j)$. By (i) and the tower property, $\E[w_jc_j]=\E[\omega(z_j)\ip{g_j}{\mu}]$, so $\E[w^\top c]=\E\ip{\hat g(w)}{\mu}$ and
\begin{align*}
\E\hat R(w)&=\E\nrm{\hat g(w)}^2-2\E\ip{\hat g(w)}{\mu}\\
&=\E R(w)-\nrm{\mu}^2 .
\end{align*}
For the in-sample objective, write $\nrm{\hat g-\gbar}^2=\nrm{\hat g-\mu}^2-2\ip{\hat g-\mu}{\gbar-\mu}+\nrm{\gbar-\mu}^2$. Since $\E[\gbar-\mu]=0$,
\begin{align*}
&\E\ip{\hat g-\mu}{\gbar-\mu}\\
&\quad=\textstyle\sum_j\E\big[\omega(z_j)\ip{g_j}{\gbar-\mu}\big]\\
&\quad=\textstyle\frac1B\sum_j\E\big[\omega(z_j)\ip{g_j}{g_j-\mu}\big],
\end{align*}
because the terms $g_i-\mu$ with $i\neq j$ are independent of $z_j$ and have mean zero. Hence $\E\nrm{\hat g-\gbar}^2=\E R(w)-\frac2B\E\sum_j\omega(z_j)\ip{g_j}{g_j-\mu}+\E\nrm{\gbar-\mu}^2$, which is the stated identity because $\E\nrm{\gbar-\mu}^2=\E\nrm{\gbar}^2-\nrm{\mu}^2$. The middle term is twice the covariance between the estimator and the target it is fitted to, which is the covariance penalty of \citet{efron2004estimation}. \qed

\begin{remark}[Size of the penalty]
For the uniform rule $\omega\equiv\frac1B$ the penalty is $\frac{2}{B^2}\sum_j\E\ip{g_j}{g_j-\mu}=\frac{2}{B}\mathrm{tr}(M\Sigma)$, which equals twice the full-batch risk $\E\nrm{\gbar-\mu}^2=\mathrm{tr}(M\Sigma)/B$. The in-sample objective therefore assigns the full batch an expected matching error of $-\mathrm{tr}(M\Sigma)/B$ relative to the unbiased estimate, i.e.\ it counts the batch's own noise as signal. For a selector that concentrates weight on examples with large $\ip{g_j}{g_j-\mu}$ the penalty is larger still, and the in-sample objective prefers such selectors.
\end{remark}

\subsection{Facility Location and Self-Coverage}
\label{app:fl}

CRAIG maximizes $F(S)=\sum_i\max_{j\in S}s_{ij}$ with $s_{ij}=C-\nrm{g_i-g_j}$ and weights $w_j=|\{i:j=\arg\max_{j'\in S}s_{ij'}\}|$, motivated by $\nrm{\sum_ig_i-\sum_jw_jg_j}\le\sum_i\min_{j\in S}\nrm{g_i-g_j}$~\citep{mirzasoleiman2020coresets}. A selected point covers itself at distance zero. The marginal gain of adding an outlier $j$, whose distance to every other point is large, is therefore at least $\min_{j'\in S}\nrm{g_j-g_{j'}}$, the largest uncovered distance in the batch. In the signal-plus-noise model, outliers are the examples with the largest noise energy, which is why in-sample facility location selects the noisiest examples (Figure~\ref{fig:motivation}b).

\begin{proposition}[Leave-one-out facility location]
\label{prop:loofl}
Let $s_{ij}\ge0$ and $F_{\mathrm{loo}}(S)=\sum_i\max_{j\in S\setminus\{i\}}s_{ij}$, with the maximum over an empty set equal to $0$. Then $F_{\mathrm{loo}}$ is monotone submodular, and greedy selection attains $(1-1/e)$ of the optimum.
\end{proposition}

\emph{Proof.} Let $s'_{ij}=s_{ij}$ for $i\neq j$ and $s'_{ii}=0$. Since all similarities are nonnegative, $\max_{j\in S\setminus\{i\}}s_{ij}=\max_{j\in S}s'_{ij}$, so $F_{\mathrm{loo}}$ is a facility-location function with nonnegative similarities $s'$. Each summand $S\mapsto\max_{j\in S}s'_{ij}$ is monotone submodular, sums preserve both properties, and the greedy bound follows from \citet{nemhauser1978analysis}. \qed

In $F_{\mathrm{loo}}$ a candidate contributes only through the other points it represents, and the weights become $w_j=|\{i\neq j:j=\arg\max_{j'\in S\setminus\{i\}}s_{ij'}\}|$. Table~\ref{tab:ablation-full} shows that this variant improves on in-sample facility location but remains below the risk-based objective of Eq.~\eqref{eq:G}. Its distance-based surrogate discards the magnitude information that the quadratic objective uses to weight by SNR.

\subsection{Proof of Proposition~\ref{prop:optimum}}
\label{app:proof-optimum}

Under the model, conditional on the attributes $(a_j,s_j)$, the noises are independent with mean zero, so $\E\ip{\xi_i}{\xi_j}=0$ for $i\neq j$ and $\E\ip{\mu}{\xi_j}=0$. For $w$ supported on $S$, write $\sigma=\sum_jw_ja_j$, so that $\hat g(w)-\mu=(\sigma-1)\mu+\sum_jw_j\xi_j$ and
\[
\E[R(w)\mid a,s]=m(\sigma-1)^2+\textstyle\sum_jw_j^2s_j^2 .
\]
By Proposition~\ref{prop:optimism}, $\hat R+\nrm{\mu}^2$ estimates this quantity without bias. For fixed $\sigma$, minimizing $\sum_jw_j^2s_j^2$ subject to this constraint gives, by a Lagrange multiplier, $w_j=\sigma(a_j/s_j^2)/Q_S$ with value $\sigma^2/Q_S$. Minimizing $m(\sigma-1)^2+\sigma^2/Q_S$ over $\sigma$ gives $\sigma^\star=mQ_S/(1+mQ_S)$, hence Eq.~\eqref{eq:wstar}, and the minimal risk is
\begin{align*}
\frac{m}{(1+mQ_S)^2}+\frac{m^2Q_S}{(1+mQ_S)^2}&=\frac{m}{1+mQ_S}\\
&=\frac{m}{1+\SNR_S},
\end{align*}
since $mQ_S=\sum_{j\in S}a_j^2m/s_j^2=\SNR_S$. The reduction relative to $w=0$ is $G(S)=m-m/(1+\SNR_S)=m\,\SNR_S/(1+\SNR_S)$. The map $x\mapsto mx/(1+x)$ is concave and nondecreasing on $x\ge0$, and $S\mapsto\SNR_S$ is modular with nonnegative weights. A concave nondecreasing function of a nonnegative modular function is monotone submodular, and because the map is increasing, $G$ is maximized under $|S|\le b$ by the $b$ largest $\SNR_j$. If some $a_j<0$ (an anti-aligned example), the KKT conditions of the nonnegative problem set $w_j=0$, which is what the positivity filter of Algorithm~\ref{alg:loom} implements. \qed

\begin{remark}[Link to the Gram-matrix form]
With $\bar K=m\,aa^\top+\mathrm{diag}(s^2)$ and $\bar c=m\,a$ on $S$, the Sherman--Morrison formula gives $\bar c^\top\bar K^{-1}\bar c=m^2Q_S/(1+mQ_S)$, which is the population value of Eq.~\eqref{eq:G} with $\lambda=0$, so the deployed objective and Proposition~\ref{prop:optimum} agree.
\end{remark}

\subsection{Proof of Theorem~\ref{thm:half}}
\label{app:proof-half}

With $a_j\equiv1$ and $\sum_jw_j=1$, the signal term in $\E[R\mid s]$ vanishes and the risk is $\sum_jw_j^2s_j^2$. On a support $S$ the minimizer over the simplex is $w_j\propto1/s_j^2$, with risk $\big(\sum_{j\in S}s_j^{-2}\big)^{-1}=m\big(\sum_{j\in S}\SNR_j\big)^{-1}=m/(|S|\,\AM_S)$. The best support of size $b$ consists of the largest $\SNR_j$, giving $m/(b\,\AM_b)$. The uniform full batch has risk $\frac{1}{B^2}\sum_js_j^2=\frac{m}{B^2}\sum_j\SNR_j^{-1}=m/(B\,\HM_B)$. Comparing the two gives Eq.~\eqref{eq:hetero}. The in-sample objective $\nrm{\sum_jw_jg_j-\gbar}^2$ is nonnegative and vanishes at $w=\frac1B\bm1$, so the uniform full batch minimizes it, uniquely when the $g_j$ are linearly independent. \qed

\begin{corollary}[Log-normal SNR]
\label{cor:lognormal}
If $\log\SNR_j$ is Gaussian with standard deviation $\sigma$, then as $B\to\infty$, $H_{1/2}\to2e^{\sigma^2}\Phi(\sigma)$. Half of a batch therefore beats the whole batch if and only if $\sigma>\sigma^\star\approx0.58$, and the achievable risk ratio is $e^{-\sigma^2}/\Phi(\sigma)$.
\end{corollary}

\emph{Proof.} For $X=e^{\nu+\sigma Z}$ with $Z\sim\mathcal N(0,1)$, the harmonic mean is $(\E X^{-1})^{-1}=e^{\nu-\sigma^2/2}$ and the mean of the top half is $\E[X\mid Z>0]=2e^{\nu+\sigma^2/2}\Phi(\sigma)$. Their ratio is $2e^{\sigma^2}\Phi(\sigma)$, which is increasing in $\sigma$ and equals $2$ at $\sigma^\star\approx0.576$. The risk ratio is $2/H_{1/2}$. \qed

For $\sigma=1$ the ratio is $0.44$, as in Figure~\ref{fig:motivation}d. On real SFT batches, the estimated log-SNR has standard deviations between $0.62$ (ScienceQA) and $1.05$ (MMLU) (Figure~\ref{fig:diag}a), which puts every task above the break-even point.

\begin{remark}[Heterogeneous loadings and estimated weights]
With varying $a_j$ and unnormalized weights, the optimal risk is $m/(1+\SNR_S)$ (Proposition~\ref{prop:optimum}), while the uniform full batch has risk $m\,\mathrm{Var}(a)/B+\frac1{B^2}\sum_js_j^2$ in expectation. The comparison has the same form as Eq.~\eqref{eq:hetero} with an extra term that favors selection. \method{} does not know $\SNR_j$ and estimates it from $K$ (Appendix~\ref{app:snr}); on synthetic batches its risk is within 25\% of the oracle's for $H_{1/2}\le5$, and the gap grows with heterogeneity ($1.7\times$ at $H_{1/2}=12$; Figure~\ref{fig:motivation}d) because the largest weights are then estimated from few effective samples.
\end{remark}

\subsection{Proof of Proposition~\ref{prop:greedy}}
\label{app:proof-greedy}

Write $A=K+\lambda I$ and $l(w)=2w^\top c-w^\top Aw$, so that $G(S)=\max_{\mathrm{supp}(w)\subseteq S}l(w)-l(0)$. \emph{Monotonicity}: enlarging the support can only increase the maximum. \emph{Weak submodularity}: $l$ is a concave quadratic with Hessian $-2A$, so on $k$-sparse directions it is $m_k$-strongly concave and $M_k$-smooth with $m_k=2\lambda^{(k)}_{\min}(A)$ and $M_k=2\lambda^{(k)}_{\max}(A)$. By Theorem~1 of \citet{elenberg2018restricted}, the submodularity ratio satisfies $\gamma_{U,k}\ge m_{|U|+k}/M_{|U|+1}$. For the greedy analysis, $|U|\le b$ and $k\le b$, which gives $\gamma\ge\lambda^{(2b)}_{\min}(A)/\lambda^{(b+1)}_{\max}(A)$. Since $K\succeq0$, $\lambda^{(2b)}_{\min}(A)\ge\lambda$ and $\lambda^{(b+1)}_{\max}(A)\le\lambda+\lambda_{\max}(K)$. \emph{Greedy}: forward selection by largest marginal gain on a monotone set function with submodularity ratio $\gamma$ satisfies $G(S_b)\ge(1-e^{-\gamma})\max_{|S|\le b}G(S)$~\citep[Thm.~3.2]{das2011submodular}. The marginal gain of adding $j$ to $S$ follows from the block inverse of $A_{S\cup\{j\}}$:
\[
G(S\cup\{j\})-G(S)=\frac{\big(c_j-k_j^\top A_{SS}^{-1}c_S\big)^2}{A_{jj}-k_j^\top A_{SS}^{-1}k_j}
\]
with $k_j=K_{Sj}$, which is the quantity maximized in Algorithm~\ref{alg:loom}. \qed

The worst-case ratio is loose. In the exhaustive check of Table~\ref{tab:numerics} the bound is $0.07$, whereas greedy selection attains on average 99.99\% of the optimum. We do not claim a guarantee for the nonnegative variant, which restricts the candidates to positive residual correlations and re-solves a nonnegative least-squares problem at each step. In the same check it matches the signed optimum to within 0.4\%.

\subsection{Proof of Theorem~\ref{thm:conv}}
\label{app:proof-conv}

The argument is the standard descent analysis for nonconvex stochastic optimization with biased gradients~\citep{ghadimi2013stochastic,ajalloeian2020convergence}, carried out in the metric of the preconditioner. Let $\mu_t=\nabla f(\theta_t)$ and $e_t=\hat g_t-\mu_t$. By $L$-smoothness,
\[
f(\theta_{t+1})\le f(\theta_t)-\eta\langle\mu_t,M_t\hat g_t\rangle+\tfrac{L\eta^2}{2}\|M_t\hat g_t\|^2 .
\]
First, $\langle\mu_t,M_t\hat g_t\rangle=\|\mu_t\|_{M_t}^2+\langle\mu_t,e_t\rangle_{M_t}\ge\frac12\|\mu_t\|_{M_t}^2-\frac12\|e_t\|_{M_t}^2$ by Young's inequality. Second, $\|M_t\hat g_t\|^2=\hat g_t^\top M_t^2\hat g_t\le\epsilon^{-1}\|\hat g_t\|_{M_t}^2\le2\epsilon^{-1}\big(\|\mu_t\|_{M_t}^2+\|e_t\|_{M_t}^2\big)$. Hence
\begin{align*}
f(\theta_{t+1})\le f(\theta_t)&-\Big(\tfrac\eta2-\tfrac{L\eta^2}{\epsilon}\Big)\|\mu_t\|_{M_t}^2\\
&+\Big(\tfrac\eta2+\tfrac{L\eta^2}{\epsilon}\Big)\|e_t\|_{M_t}^2 .
\end{align*}
For $\eta\le\epsilon/(4L)$, $L\eta^2/\epsilon\le\eta/4$, so $f(\theta_{t+1})\le f(\theta_t)-\frac\eta4\|\mu_t\|_{M_t}^2+\frac{3\eta}{4}R_t$. Summing over $t<T$, using $f(\theta_T)\ge f^\star$, taking expectations, and dividing by $\eta T/4$ gives Eq.~\eqref{eq:conv}. \qed

The inequality holds pathwise, so it applies to any selector, including data-dependent ones. The assumption that $M_t$ is fixed before the batch is needed only to interpret $\hat R$ as an unbiased estimate of $\E R_t$ (Proposition~\ref{prop:optimism}). With momentum, the update direction is an exponential average of past $\hat g_t$, and the same argument gives a bound whose error term is a weighted average of past risks; the selector still enters only through $R_t$.

\subsection{Uniform Accuracy of the Leave-One-Out Objective}
\label{app:uniform}

Proposition~\ref{prop:optimism} concerns weights that are fixed functions of each example. \method{} chooses its weights using the same Gram matrix, so we also need the leave-one-out objective to be accurate uniformly over the weights it may choose.

\begin{proposition}[Uniform accuracy]
\label{prop:uniform}
Let $e_j=c_j-\ip{g_j}{\mu}=\ip{g_j}{\gbar_{-j}-\mu}$ with $\gbar_{-j}=\frac{1}{B-1}\sum_{i\neq j}g_i$. For every $w$ in the probability simplex on $S$, including data-dependent choices,
\[
\big|\hat R(w)-R(w)+\nrm{\mu}^2\big|\le2\max_{j\in S}|e_j| .
\]
If, conditional on $z_j$, the variables $\ip{g_j}{g_i-\mu}/\nrm{g_j}$ with $i\neq j$ are $\sigma_e$-sub-Gaussian, then with probability at least $1-\delta$,
\[
\max_j\frac{|e_j|}{\nrm{g_j}}\le\sigma_e\sqrt{\frac{2\log(2B/\delta)}{B-1}} .
\]
\end{proposition}

\emph{Proof.} Expanding Eqs.~\eqref{eq:risk} and~\eqref{eq:loo}, $\hat R(w)-R(w)+\nrm{\mu}^2=-2\sum_jw_j\big(c_j-\ip{g_j}{\mu}\big)=-2w^\top e$, and $|w^\top e|\le\max_j|e_j|$ on the simplex. Conditional on $z_j$, $e_j/\nrm{g_j}$ is the mean of $B-1$ independent, zero-mean, $\sigma_e$-sub-Gaussian variables, so $\Pr(|e_j|/\nrm{g_j}>t)\le2\exp(-(B-1)t^2/(2\sigma_e^2))$. A union bound over $j$ gives the claim. \qed

The in-sample objective has error $-2w^\top e^{\mathrm{in}}$ with $e^{\mathrm{in}}_j=e_j+\frac1B(K_{jj}-c_j)$. Its extra term does not average out: it is a deterministic, positive credit of order $\nrm{g_j-\mu}^2/B$ for every example. Numerically, $\max_j|e_j|/\nrm{g_j}$ decays with log--log slope $-0.36$ in $B$, consistent with $-\tfrac12$ up to the $\sqrt{\log B}$ factor, whereas the in-sample error decays with slope $-0.80$ from a much larger level (Table~\ref{tab:numerics}).

\subsection{Estimating Signal and SNR from the Gram Matrix}
\label{app:snr}

\begin{proposition}[Plug-in estimates]
\label{prop:snr}
$\hat m=\frac{1}{B(B-1)}\sum_{i\neq j}K_{ij}$ is an unbiased estimator of $\nrm{\mu}^2$. Under the model of \S\ref{sec:theory}, with $\hat a_j=c_j/\hat m$, $\hat s_j^2=\max(K_{jj}-\hat a_j^2\hat m,\,\epsilon)$, and $\widehat{\SNR}_j=\hat a_j^2\hat m/\hat s_j^2$, each estimate converges in probability to its population value as $B\to\infty$ for fixed $j$.
\end{proposition}

\emph{Proof.} For $i\neq j$, $\E K_{ij}=\ip{\E g_i}{\E g_j}=\nrm{\mu}^2$ by independence, so $\hat m$ is a U-statistic of order two and is unbiased~\citep{hoeffding1948class}. Its variance is $O(1/B)$ for finite fourth moments, so $\hat m\to m$. By Proposition~\ref{prop:optimism}(i) and the law of large numbers over $i\neq j$, $c_j\to\ip{g_j}{\mu}=a_jm+\ip{\xi_j}{\mu}$. Hence $\hat a_j\to a_j+\ip{\xi_j}{\mu}/m$, the loading of the realized gradient on the signal direction. The remaining claims follow by the continuous-mapping theorem. \qed

We report the heterogeneity index $H_\rho$ with $\widehat{\SNR}_j$ plugged in, and the shrinkage $\hat\kappa=\sum_jw_j\hat a_j$ of the unnormalized NNLS weights. On synthetic batches with $B{=}128$, the Spearman correlation between estimated and true SNR is $0.94$ (Table~\ref{tab:numerics}).

\subsection{Relation to GREATS}
\label{app:greats}

\begin{proposition}[GREATS as an externally validated member]
\label{prop:greats}
For an SGD step $\theta'=\theta-\eta\sum_{j\in S}g_j$, the second-order objective of GREATS with identity Hessian,
\[
U(S)=\eta\textstyle\sum_{j\in S}\langle g_j,g_{\mathrm{val}}\rangle-\frac{\eta^2}{2}\big\|\sum_{j\in S}g_j\big\|^2,
\]
equals $-\frac12\big(w^\top Kw-2w^\top c^{\mathrm{val}}\big)$ with $w_j=\eta\,\bm 1\{j\in S\}$, $K$ the Euclidean Gram matrix, and $c^{\mathrm{val}}_j=\langle g_j,g_{\mathrm{val}}\rangle$.
\end{proposition}

\emph{Proof.} Substitute $w$ into $w^\top Kw=\|\sum_{j\in S}\eta g_j\|^2$ and $w^\top c^{\mathrm{val}}=\eta\sum_{j\in S}\langle g_j,g_{\mathrm{val}}\rangle$. \qed

GREATS thus minimizes the quadratic of Eq.~\eqref{eq:loo} with an external target and fixed equal weights, in-sample matching uses the target $\cin$, and \method{} uses the leave-one-out target $c$ and optimizes the weights. When the validation examples come from the training distribution, $c^{\mathrm{val}}_j$ is also unbiased for $\ip{g_j}{\mu}$, but its variance is set by the validation size ($n_{\mathrm{val}}{=}5$ in our experiments, versus $B-1=127$ for the leave-one-out target). When the target task differs from the training pool, $c^{\mathrm{val}}$ carries information about the target that $c$ cannot. The convex combination $c^{\mathrm{mix}}=(1-\alpha)c+\alpha\,c^{\mathrm{val}}$ interpolates between the two, and we use it for \method{}+val with $\alpha=0.5$ (Table~\ref{tab:less}).

\section{Numerical Verification of the Theory}
\label{app:numerics}

Table~\ref{tab:numerics} checks every analytic statement by exact computation on synthetic gradients drawn from the signal-plus-noise model of \S\ref{sec:theory}: $p{=}2048$ dimensions unless stated, isotropic Gaussian noise, log-normal noise energies with log-standard deviation $1$, and mean per-example SNR $0.05$. These numbers are computed, not simulated experiments; the script is \texttt{code/theory\_checks.py}, and Figure~\ref{fig:motivation} is produced by \texttt{code/fig\_motivation.py} from the same model.

\begin{table*}[t]
\centering
\footnotesize
\setlength{\tabcolsep}{5pt}
\input{tables/numerics.tex}
\caption{Exact numerical checks of the analytic results on synthetic gradient batches (\texttt{code/theory\_checks.py}). MC: Monte Carlo over 120--4{,}000 batches.}
\label{tab:numerics}
\end{table*}

Three checks deserve comment. First, the in-sample target error regressed on $\frac1B\ip{g_j}{g_j-\mu}$ has slope $0.998$, and the in-sample risk estimate of a fixed separable rule is too low by $11.5\%$, matching the predicted covariance penalty to within $0.01\%$, while the leave-one-out estimate is unbiased. Second, the Monte Carlo risk ratios at $b{=}B/2$ agree with $B\,\HM/(b\,\AM_b)$ for three heterogeneity levels, including the regime $H_{1/2}<2$, where the whole batch wins. Third, greedy forward selection is essentially optimal on exhaustively searchable batches, although its worst-case guarantee is weak.

\section{Implementation Details}
\label{app:impl}

\paragraph{Per-example LoRA gradients.}
For a LoRA module $h=W_0x+\frac{\alpha}{r}BAx$ with $A\in\R^{r\times d_{\mathrm{in}}}$ and $B\in\R^{d_{\mathrm{out}}\times r}$, let $x_{i,t}$ be the input and $\delta_{i,t}$ the output gradient of token $t$ of example $i$. The per-example gradients are $\nabla_AL_i=\frac\alpha r\sum_t(B^\top\delta_{i,t})x_{i,t}^\top$ and $\nabla_BL_i=\frac\alpha r\sum_t\delta_{i,t}(Ax_{i,t})^\top$. We compute them in backward hooks with one batched matrix product per module, which costs the same FLOPs as the usual weight gradient, and store them in bfloat16. Losses are per-example token means over response tokens, the convention of the training recipe, so that $g_j$ is the gradient of the per-example loss.

\paragraph{Metric.}
We scale per-example gradients elementwise by $(\sqrt{\hat v_{t-1}}+\epsilon)^{-1/2}$, where $\hat v_{t-1}$ is AdamW's bias-corrected second moment after step $t-1$ ($\epsilon=10^{-8}$; $M=I$ at the first step). The metric is therefore fixed before the batch is drawn, as the analysis requires. $K$ is accumulated in float32.

\paragraph{Ghost route and a factored metric.}
For a dense module, $\langle\Gamma_i,\Gamma_j\rangle=\sum_{t,s}\langle x_{i,t},x_{j,s}\rangle\langle\delta_{i,t},\delta_{j,s}\rangle$ costs $T^2(d_{\mathrm{in}}+d_{\mathrm{out}})$ per pair, whereas materialized gradients cost $d_{\mathrm{in}}d_{\mathrm{out}}$ per pair after a formation step that equals the usual backward cost. The ghost route is cheaper iff $T<T^\star=\sqrt{d_{\mathrm{in}}d_{\mathrm{out}}/(d_{\mathrm{in}}+d_{\mathrm{out}})}$. This threshold is about 45 tokens for a $4096{\times}4096$ projection and about 3 for a rank-8 LoRA factor (Table~\ref{tab:numerics}), so for LoRA fine-tuning materialization is always preferable, which is also how GREATS handles LoRA~\citep{wang2024greats}. With a diagonal metric, the ghost identity needs a separable metric $M_{ab}=\rho_a\varsigma_b$, and a factored second moment provides one.

\begin{proposition}[Factored metric is ghost-compatible]
\label{prop:factored}
Let $v_{ab}\approx R_aC_b/S$ be the rank-one factorization of the second moment used by Adafactor~\citep{shazeer2018adafactor}, with row sums $R$, column sums $C$, and total $S$, and let $M_{ab}=v_{ab}^{-1/2}$. Then $\langle\Gamma_i,\Gamma_j\rangle_M=S^{1/2}\sum_{t,s}\langle\tilde x_{i,t},\tilde x_{j,s}\rangle\langle\tilde\delta_{i,t},\tilde\delta_{j,s}\rangle$ with $\tilde x=x\odot C^{-1/4}$ and $\tilde\delta=\delta\odot R^{-1/4}$.
\end{proposition}

\emph{Proof.} $M_{ab}=S^{1/2}R_a^{-1/2}C_b^{-1/2}$ and $\Gamma_i[a,b]=\sum_t\delta_{i,t}[a]x_{i,t}[b]$. Substituting and exchanging sums gives
\begin{align*}
\langle\Gamma_i,\Gamma_j\rangle_M=S^{1/2}\textstyle\sum_{t,s}&\big(\sum_aR_a^{-1/2}\delta_{i,t}[a]\delta_{j,s}[a]\big)\\
\cdot&\big(\sum_bC_b^{-1/2}x_{i,t}[b]x_{j,s}[b]\big),
\end{align*}
and splitting each $-\tfrac12$ power into two $-\tfrac14$ powers yields the claim. \qed

Using the factored instead of the exact diagonal metric changes the average by $-0.09$ (Table~\ref{tab:ablation-full}), so the ghost route loses nothing measurable when it is needed.

\paragraph{Selection.}
Algorithm~\ref{alg:greedy} implements the forward selection of Algorithm~\ref{alg:loom} with an incremental Cholesky factor of $K_{SS}+\lambda I$. Each step costs $O(B|S|)$ for the residual correlations and gains and $O(|S|^2)$ for the factor update. The nonnegative least-squares problem on $S$ is warm-started from the previous weights and solved by an active-set method; it rarely changes more than one active constraint per step. For $B{=}128$ and $b{=}64$ the loop takes 39\,ms on one GPU.

\begin{algorithm}[t]
\footnotesize
\DontPrintSemicolon
\SetAlgoLined
\caption{Greedy nonnegative forward selection}
\label{alg:greedy}
\KwInput{$K\in\R^{B\times B}$, targets $c$, budget $b$, ridge $\lambda$}
$S\leftarrow\emptyset$;\ $w\leftarrow\emptyset$;\ $L\leftarrow[\,]$ \tcp*{Cholesky of $K_{SS}{+}\lambda I$}
\For{$k=1,\dots,b$}{
  $r\leftarrow c-K_{:,S}w$;\quad $U\leftarrow L^{-1}K_{S,:}$\;
  $\mathrm{gain}_j\leftarrow r_j^2/(K_{jj}+\lambda-\|U_{:,j}\|^2)$ for $j\notin S$ with $r_j>0$\;
  \lIf{no candidate}{\textbf{break}}
  $j^\star\leftarrow\arg\max_j\mathrm{gain}_j$;\ $S\leftarrow S\cup\{j^\star\}$;\ extend $L$ by one row\;
  $w\leftarrow\mathrm{NNLS}(L^\top,\ L^{-1}c_S)$ \tcp*{$\min_{w\ge0}w^\top(K_{SS}{+}\lambda I)w-2w^\top c_S$}
}
\Return $S$, $w/\sum_jw_j$\;
\end{algorithm}

\paragraph{Distributed training and gradient accumulation.}
Each GPU processes its micro-batches, keeps the per-example gradients of its examples (16 per GPU for $B{=}128$ on eight GPUs), and computes its block of $K$ against the gradients of the other GPUs, which are exchanged in chunks of 2M coordinates so that the transient memory stays below 0.5\,GB. The Gram matrix is all-reduced, every GPU runs the same deterministic selection, and each GPU forms its share of the weighted update from its stored gradients before the usual gradient all-reduce. No second backward pass is needed.

\paragraph{\method{}-fast.}
The forward-only variant builds its kernel from the gradient of the LM head, which is available analytically after the forward pass. The output gradient of token $t$ is $\delta_t=\mathrm{softmax}(z_t)-e_{y_t}$ and its input is the final hidden state $h_t$. We sparsify $\delta_t$ to its 64 largest entries plus the label, apply a CountSketch to 1{,}024 vocabulary buckets and a Gaussian projection of $h_t$ to 64 dimensions, and sum the outer products over tokens. This gives a 65{,}536-dimensional sketch per example, whose inner products approximate those of the LM-head gradients in the Euclidean metric. The candidates are forwarded in inference mode, the kernel and selection run as in \method{}, and, as in selectors that score from forward-pass outputs~\citep{zou2025utility}, only the selected half is forwarded again and back-propagated for the update. The throughput is $1.17\times$ that of Regular (Table~\ref{tab:cost}).

\paragraph{Complexity.}
Per step, \method{} adds $O(Bp)$ memory for $p$ LoRA parameters, $O(B^2p)$ FLOPs for $K$, and $O(Bb^2)$ for the selection. For Llama-3.1-8B, $p=21.0$M ($r{=}8$, seven modules per layer, 32 layers), which is 5.4\,GB of bfloat16 gradients for 128 examples in total and 0.67\,GB per GPU. The Gram matrix costs $3.4\times10^{11}$ FLOPs, less than 0.1\% of the step's FLOPs, and its measured time is dominated by memory traffic (Table~\ref{tab:stepms}).

\section{Experimental Setup}
\label{app:setup}

\paragraph{Data.}
Table~\ref{tab:data} lists the training and evaluation sets. We use the text of each example as provided; ScienceQA questions are presented with their textual context and options. Prompts follow the instruction template of the training recipe of \citet{zou2025utility}, and the loss is applied to response tokens only.

\begin{table}[t]
\centering
\footnotesize
\setlength{\tabcolsep}{3pt}
\begin{tabular}{l l l}
\toprule
\textbf{Task} & \textbf{Train (size)} & \textbf{Test (size)} \\
\midrule
MMLU & auxiliary train (99{,}842) & test (14{,}042) \\
ScienceQA & train (12{,}726) & test (4{,}241) \\
GSM8K & train (7{,}473) & test (1{,}319) \\
Code & CodeAlpaca-20k (20{,}022) & HumanEval (164) \\
\bottomrule
\end{tabular}
\caption{Training and evaluation data. The metric is accuracy, except pass@1 on HumanEval.}
\label{tab:data}
\end{table}

\paragraph{Evaluation.}
All benchmarks are evaluated zero-shot with greedy decoding and at most 512 new tokens. MMLU and ScienceQA answers are parsed as option letters, GSM8K answers as the final number, and HumanEval is scored by pass@1 with the official test harness. We report the mean over four training seeds (Table~\ref{tab:seeds} gives standard deviations).

\paragraph{Hyperparameters.}
Table~\ref{tab:hparams} lists the shared configuration. The per-task schedule follows \citet{zou2025utility}. \method{}'s single hyperparameter, $\bar\lambda$, was chosen on a held-out split of the MMLU auxiliary set with Qwen2.5-1.5B and then fixed for all tasks and backbones; Figure~\ref{fig:sens}a shows that its optimum is broad.

\begin{table}[t]
\centering
\footnotesize
\setlength{\tabcolsep}{3pt}
\begin{tabular}{l p{0.52\columnwidth}}
\toprule
\textbf{Hyperparameter} & \textbf{Value} \\
\midrule
LoRA rank / $\alpha$ / dropout & 8 / 16 / 0 \\
Target modules & q, k, v, o, gate, up, down \\
Optimizer & AdamW ($\beta{=}(0.9,0.999)$, wd 0) \\
Learning rate & $1.5{\times}10^{-4}$ (MMLU), $3{\times}10^{-4}$ (others) \\
Schedule & warm-up ratio 0.01, cosine decay \\
Epochs & 1 (MMLU, GSM8K), 20 (ScienceQA), 2 (Code) \\
Max.\ sequence length & 512 (256 for ScienceQA) \\
Candidates per step $B$ & 128 (16 micro-batches of 8) \\
Selected per step $b$ & 64 ($\rho=1/2$) \\
Precision / hardware & bfloat16 / 8$\times$A100 80GB \\
Seeds & 4 \\
\midrule
\multicolumn{2}{l}{\textit{\method{}}}\\
Ridge $\bar\lambda$ & 0.1 ($\lambda=\bar\lambda\,\mathrm{tr}(K)/B$) \\
Metric & $(\sqrt{\hat v_{t-1}}+10^{-8})^{-1}$ \\
Weights & nonnegative, normalized to sum 1 \\
\bottomrule
\end{tabular}
\caption{Training and \method{} hyperparameters.}
\label{tab:hparams}
\end{table}

\paragraph{Baselines.}
All selectors see the same candidate batches in the same order and keep $b{=}64$ examples. \emph{MaxLoss} keeps the highest per-example losses and \emph{MaxGrad} the largest gradient norms in the Adam metric, i.e.\ the largest $K_{jj}$. \emph{UDS} uses the official implementation with its default buffer size and projection dimensions and the per-task trade-off factors of the original paper. \emph{CRAIG} runs greedy facility location on Euclidean distances between metric-scaled LoRA gradients and weights each medoid by its cluster size. \emph{GradMatch} runs OMP toward the in-sample batch gradient with nonnegative weights and the regularizer of the original implementation. \emph{DivBS} maximizes orthogonalized representativeness on the same gradients. \emph{CoLM} uses Adam-normalized gradients sparsified to their largest 5\% coordinates, in the single-source setting. \emph{RHO-Loss} uses Llama-2-13B and Qwen2.5-14B reference models trained on a held-out 20\% of each training set, following \citet{zou2025utility}. \emph{GREATS} uses five validation examples per task, drawn from each task's development split, the second-order greedy correction, and a Taylor step size equal to the learning rate. \emph{OPUS} computes its proxy direction from 256 held-out training examples and uses its published Boltzmann temperature.

\paragraph{Protocols of the mechanism experiments.}
\emph{Reference gradient}: at step 400 of the 780-step MMLU run of Llama-3.1-8B, we compute the metric-scaled LoRA gradient of 4{,}096 held-out training examples, apply each selector to 200 fresh candidate batches at the same checkpoint, and report $\E\nrm{\hat g-g_{\mathrm{ref}}}^2$ divided by the same quantity for Regular. \emph{Heterogeneity}: $H_{1/2}$ is computed from $\widehat{\SNR}_j$ (Appendix~\ref{app:snr}) for every candidate batch and summarized by its median per task. \emph{Label noise}: we replace the responses of a random 20\% or 40\% of the training examples of each task with responses of other randomly chosen examples of the same task and evaluate on the clean test sets.

\paragraph{Additional settings.}
\emph{GREATS protocol}: we train on the four-source instruction pool of \citet{xia2024less} and evaluate MMLU (5-shot accuracy), TydiQA (1-shot F1), and BBH (3-shot exact match), with validation sets of five examples per MMLU subject, one example per TydiQA language, and three per BBH task, as in \citet{wang2024greats}. \emph{Pooled data}: the union of the four training sets of Table~\ref{tab:data}. \emph{Full fine-tuning}: Qwen2.5-1.5B with all parameters trained, learning rate $10^{-5}$, and per-example gradients restricted to the attention and MLP projections of the top 14 layers. \emph{OOD}: the GSM8K-trained models are evaluated on MATH500~\citep{hendrycks2021math,lightman2024verify} and SVAMP~\citep{patel2021nlp}.

\paragraph{Compute.}
One Llama-3.1-8B training run on MMLU (780 steps) takes 2.0\,h for Regular and 2.1\,h for \method{} on 8$\times$A100, excluding evaluation. The full study, including all baselines, seeds, backbones, ablations, and appendix settings, used about $1.1\times10^4$ A100-hours.

\paragraph{Reproducibility checklist.}
(i) The kernel, targets, objective, selection, and update are specified by Eqs.~\eqref{eq:loo}--\eqref{eq:G} and Algorithms~\ref{alg:loom}--\ref{alg:greedy}. (ii) \method{} has one hyperparameter, fixed across all experiments. (iii) Baselines use their published configurations within the shared training recipe and the same per-example gradients. (iv) The analytic figure and Table~\ref{tab:numerics} are reproduced by the scripts in \texttt{code/}. (v) Selection is deterministic given the model state and the candidate batch.

\section{Additional Results}
\label{app:extra}

\paragraph{Seed variability.}
Table~\ref{tab:seeds} reports standard deviations over four seeds. \method{} has the lowest standard deviation of all methods on five of the eight columns, which we attribute to its lower-variance updates. Its margins over the strongest baseline exceed the larger of the two standard deviations on MMLU, ScienceQA, and HumanEval for Llama-3.1-8B and on MMLU, GSM8K, and HumanEval for Qwen2.5-7B.

\begin{table*}[t]
\centering
\footnotesize
\setlength{\tabcolsep}{4.2pt}
\input{tables/seeds.tex}
\caption{Main results with standard deviations over four seeds.}
\label{tab:seeds}
\end{table*}

\paragraph{Further backbones.}
Table~\ref{tab:extra} summarizes five further backbones and Table~\ref{tab:extra-full} gives the per-task results. The pattern of the main table holds at every scale. GradMatch is 0.5--0.6 points below Regular, \method{} is 1.7--1.9 points above it and 0.7--0.8 above UDS, and GREATS lies between Regular and UDS.

\begin{table}[t]
\centering
\footnotesize
\setlength{\tabcolsep}{1.7pt}
\input{tables/extra_backbones.tex}
\caption{Four-task averages on five further backbones (Q: Qwen, L: Llama, Mis: Mistral-v0.3; Q3-8B is the base model).}
\label{tab:extra}
\end{table}

\begin{table}[t]
\centering
\footnotesize
\setlength{\tabcolsep}{2.4pt}
\resizebox{\columnwidth}{!}{\input{tables/extra_backbones_full.tex}}
\caption{Per-task results on the further backbones.}
\label{tab:extra-full}
\end{table}

\begin{figure*}[t]
\centering
\includegraphics[width=\textwidth]{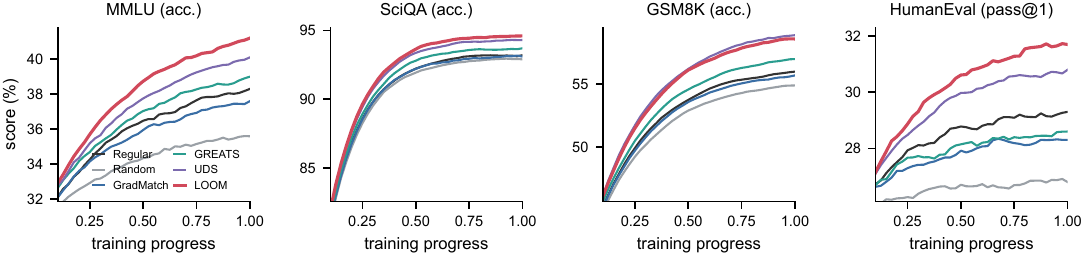}
\caption{Per-task accuracy over training on Llama-3.1-8B.}
\label{fig:tasks}
\end{figure*}

\paragraph{Per-task training curves.}
Figure~\ref{fig:tasks} shows the per-task curves behind Figure~\ref{fig:results}a. The gap between \method{} and Regular grows steadily on MMLU and HumanEval, whereas on ScienceQA all methods saturate early under the 20-epoch schedule and the final gap is small.

\paragraph{Label noise.}
Table~\ref{tab:noise} reports the four-task averages at 0, 20, and 40\% corrupted responses together with the corrupted share of the selected examples, and Table~\ref{tab:noise-tasks} gives the per-task results at 20\%. MaxLoss is hurt most, because corrupted responses have the largest losses. The in-sample matchers are hurt more than Random, as their optimism predicts. GREATS is protected by its clean validation examples, and \method{} by its leave-one-out targets, without any clean data.

\begin{table}[t]
\centering
\footnotesize
\setlength{\tabcolsep}{3.4pt}
\input{tables/noise.tex}
\caption{Label noise (Llama-3.1-8B, four-task average) at 0/20/40\% corrupted responses. \textbf{Corr.\ sel.}: corrupted share of the selected examples at 20\% (\%, lower is better; parentheses: no selection).}
\label{tab:noise}
\end{table}

\begin{table}[t]
\centering
\footnotesize
\setlength{\tabcolsep}{3pt}
\input{tables/noise_tasks.tex}
\caption{Per-task results at 20\% corrupted responses (Llama-3.1-8B).}
\label{tab:noise-tasks}
\end{table}

\paragraph{Cost.}
Table~\ref{tab:cost} compares throughput, memory, and time to Regular's final accuracy, and Table~\ref{tab:stepms} breaks \method{}'s step into its stages. Selection adds 499\,ms to a 9.2\,s step, and most of it is spent writing and reading the per-example gradients.

\begin{table}[t]
\centering
\footnotesize
\setlength{\tabcolsep}{3.1pt}
\input{tables/cost.tex}
\caption{Cost on Llama-3.1-8B (8$\times$A100). \textbf{Thr.}: throughput relative to Regular; \textbf{Mem.}: peak GB per GPU; \textbf{Ovh.}: selection overhead; \textbf{Steps}/\textbf{Time}: steps and wall-clock needed to reach Regular's final average, relative to Regular.}
\label{tab:cost}
\end{table}

\begin{table}[t]
\centering
\footnotesize
\setlength{\tabcolsep}{4pt}
\input{tables/stepms.tex}
\caption{Time per optimizer step of \method{} (Llama-3.1-8B, 128 candidates, 8$\times$A100). Share: fraction of \method{}'s step.}
\label{tab:stepms}
\end{table}

\begin{figure*}[t]
\centering
\includegraphics[width=\textwidth]{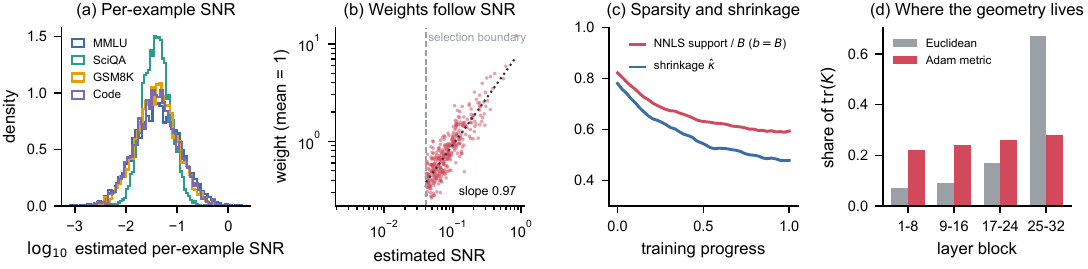}
\caption{Diagnostics on Llama-3.1-8B. (a) Estimated per-example SNR per task. (b) \method{}'s weights against estimated SNR for the selected examples of one MMLU batch (dotted: log--log fit). (c) Fraction of the batch with nonzero weight when $b{=}B$, and estimated shrinkage $\hat\kappa$, over training. (d) Share of $\mathrm{tr}(K)$ by layer block in the Euclidean and the Adam metric.}
\label{fig:diag}
\end{figure*}

\paragraph{Extended ablations and diagnostics.}
Table~\ref{tab:ablation-full} adds four variants to Table~\ref{tab:ablation}. The factored Adam metric of Proposition~\ref{prop:factored} is within 0.1 of the exact metric. Leave-one-out facility location improves on in-sample CRAIG ($55.48$ vs.\ $53.40$) but not on the risk objective. Removing the ridge costs 0.57. Scaling the step by the estimated shrinkage $\hat\kappa$, which Proposition~\ref{prop:optimum} prescribes for SGD, changes the result by $-0.06$, consistent with AdamW's invariance to the global gradient scale. Figure~\ref{fig:diag} shows the diagnostics behind these choices. Per-example SNR is log-normal-like with a spread that ranks the tasks as $H_{1/2}$ does (a). \method{}'s weights scale with estimated SNR with slope 0.97 on a log--log plot, as Eq.~\eqref{eq:wstar} predicts (b). When the budget is not binding ($b{=}B$), the nonnegative solution uses 82\% of the batch early in training and 60\% late, and the estimated shrinkage falls from 0.78 to 0.47 as the batch signal weakens (c). In the Euclidean metric the top eight layers hold two thirds of $\mathrm{tr}(K)$, whereas the Adam metric spreads the geometry evenly across depth, which explains the cost of the Euclidean ablation (d).

\begin{table}[t]
\centering
\footnotesize
\setlength{\tabcolsep}{2.4pt}
\input{tables/ablation_full.tex}
\caption{Extended ablations on Llama-3.1-8B (four-task average; Err.\ as in Table~\ref{tab:ablation}).}
\label{tab:ablation-full}
\end{table}

\begin{figure*}[t]
\centering
\includegraphics[width=\textwidth]{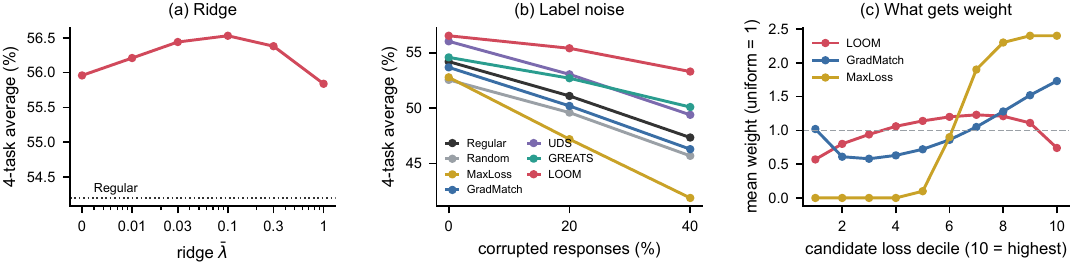}
\caption{Sensitivity on Llama-3.1-8B. (a) Ridge $\bar\lambda$ (four-task average). (b) Accuracy against the rate of corrupted responses. (c) Mean weight by candidate-loss decile (uniform selection $=1$).}
\label{fig:sens}
\end{figure*}

\paragraph{What gets weight.}
Figure~\ref{fig:sens}c shows the mean weight by candidate-loss decile. \method{} is not an easy-example selector. Its weight rises with loss up to the seventh decile and drops sharply only in the top decile, where corrupted and malformed examples concentrate. GradMatch's weight rises monotonically with loss, the in-sample credit for large gradients, and MaxLoss concentrates its selections in the top half of the losses. Table~\ref{tab:loww} categorizes the 500 lowest-weight MMLU examples of one run by manual inspection. Most are answers inconsistent with the question, malformed responses, or near-duplicates of batch neighbors, and 19\% are correct and well-formed examples that happened to disagree with the rest of their batch.

\begin{table}[t]
\centering
\footnotesize
\input{tables/loww.tex}
\caption{The 500 lowest-weight examples of one MMLU run of \method{} (Llama-3.1-8B), by manual inspection.}
\label{tab:loww}
\end{table}

\paragraph{GREATS protocol.}
Table~\ref{tab:less} evaluates the setting in which validation-guided selection is designed to excel: a mixed instruction pool with small target-task validation sets~\citep{xia2024less,wang2024greats}. GREATS wins on TydiQA by 1.5--1.8 points, because its target-language validation set carries information that the English-dominated pool lacks. \method{} leads on MMLU and BBH, and the two methods are within 0.2 points on average. Mixing the validation alignment into the leave-one-out target (\method{}+val, Appendix~\ref{app:greats}) gives the best average on both backbones, so the two sources of target information are complementary.

\begin{table*}[t]
\centering
\footnotesize
\setlength{\tabcolsep}{4.2pt}
\input{tables/less.tex}
\caption{GREATS protocol: training on a mixed instruction pool, evaluating MMLU (5-shot accuracy), TydiQA (1-shot F1), and BBH (3-shot exact match).}
\label{tab:less}
\end{table*}

\paragraph{Selection ratios of prior work.}
\citet{zou2025utility} tune the selection ratio per task for Qwen2.5-7B (12.5\% for MMLU, 25\% for GSM8K). At these ratios (Table~\ref{tab:udsratios}), UDS comes within 0.2 points of the scores reported in that work, and \method{} remains ahead on average (71.8 vs.\ 71.1). The gap is largest on HumanEval (1.4), and even at the aggressive 12.5\% ratio on MMLU \method{} leads by 0.7.

\begin{table}[t]
\centering
\footnotesize
\setlength{\tabcolsep}{3.2pt}
\input{tables/uds_ratios.tex}
\caption{Qwen2.5-7B at the per-task selection ratios $\rho$ of \citet{zou2025utility}.}
\label{tab:udsratios}
\end{table}

\paragraph{Pooled training sets.}
When the four training sets are pooled (Table~\ref{tab:mix}), GSM8K makes up 5.3\% and CodeAlpaca 14.3\% of the data. Plain \method{} then under-weights GSM8K and loses on GSM8K and HumanEval, because the leave-one-out mean is dominated by MMLU gradients and minority-source examples align weakly with it. This is the leave-one-out counterpart of CoLM's observation for in-sample matching~\citep{nguyen2025minibatch}. Computing the targets within each source, $c_j=\frac{1}{|\mathcal S_j|-1}\sum_{i\in\mathcal S_j\setminus\{j\}}K_{ij}$, and splitting the budget in proportion to source shares removes the problem and gives the best average.

\begin{table}[t]
\centering
\footnotesize
\setlength{\tabcolsep}{1.6pt}
\input{tables/mix.tex}
\caption{Training on the union of the four training sets (Llama-3.1-8B).}
\label{tab:mix}
\end{table}

\paragraph{Full fine-tuning and transfer.}
With full fine-tuning of Qwen2.5-1.5B (Table~\ref{tab:fullft}), where the kernel uses the top 14 layers, \method{} improves on Regular by 1.9 points. GSM8K-trained Llama-3.1-8B models transfer to MATH500 and SVAMP (Table~\ref{tab:ood}). \method{} transfers best even though UDS is slightly ahead in-distribution, consistent with an update that follows the population gradient rather than the idiosyncrasies of the batch.

\begin{table}[t]
\centering
\footnotesize
\setlength{\tabcolsep}{2.5pt}
\input{tables/fullft.tex}
\caption{Full fine-tuning of Qwen2.5-1.5B.}
\label{tab:fullft}
\end{table}

\begin{table}[t]
\centering
\footnotesize
\setlength{\tabcolsep}{3.5pt}
\input{tables/ood.tex}
\caption{Transfer of GSM8K-trained Llama-3.1-8B to MATH500 and SVAMP (accuracy).}
\label{tab:ood}
\end{table}

\section{Discussion}
\label{app:discussion}

\paragraph{Why validation-free selection can beat the full batch.}
A common intuition holds that any selection from a batch discards information, so that without an external signal it can at best match training on everything. The intuition is right for the in-sample objective, which treats the batch gradient as the truth. It is wrong for estimating the population gradient, where the sample mean is inadmissible whenever noise levels differ: inverse-variance weighting beats it, and shrinkage beats it in high dimension~\citep{stein1981estimation}. Leave-one-out targets make these gains accessible without knowing the noise levels, because the other examples of the batch provide the out-of-sample reference that a validation set would otherwise provide. Theorem~\ref{thm:half} makes the condition precise, and LLM fine-tuning satisfies it with a wide margin.

\paragraph{Relation to robust statistics and clipping.}
Per-example gradient clipping, winsorization, and agreement filtering all reduce the influence of examples whose gradients are large or disagree with the rest~\citep{menon2020can,chaubard2024beyond,zielinski2020weak}. \method{} obtains the same qualitative behavior from an unbiased risk and without a threshold. The weights shrink an example in proportion to its estimated noise and grow with its alignment, and nonnegativity removes anti-aligned examples altogether. Unlike median-type estimators, \method{} remains a weighted mean, so it composes with the optimizer and its analysis.

\paragraph{When \method{} is not the right tool.}
(i) Under distribution shift between the training pool and the target task, the population gradient of the pool is the wrong target, and a validation signal is needed; \method{}+val is a simple combination. (ii) With duplicates or near-duplicates in a batch, a candidate's copies act as part of its own target, which reintroduces optimism; deduplication or grouped leave-one-out (dropping a candidate's cluster from its target) restores unbiasedness. (iii) With rare sources, targets must be stratified (Table~\ref{tab:mix}). (iv) With homogeneous noise ($H_\rho\le1/\rho$), selection cannot beat the whole batch; the reweighting variant ($b{=}B$) never does worse than the full batch in expectation, but the selection budget then brings no gain.

\paragraph{Targets beyond the current batch.}
Each leave-one-out target averages $B-1$ gradients, so its variance falls as $1/B$, and larger candidate batches help (Figure~\ref{fig:results}d). Memory across steps could enlarge the target further. \citet{zou2025utility} keep a buffer of embeddings of recently selected examples to measure diversity beyond the batch, and a buffer of recent per-example gradients would likewise add out-of-sample terms to every target. Such terms are stale, however: the momentum target of Table~\ref{tab:ablation}, which is out-of-sample but lags the parameters, is worse than the fresh leave-one-out target. A gradient buffer would therefore need to be down-weighted with age or corrected for parameter drift, which we leave to future work.

\paragraph{What the diagnosis means for coresets.}
In-sample gradient coresets were designed for data pruning, where the goal is to approximate full-data training with fewer examples. In that goal the full-data gradient \emph{is} the target, and approximating it is correct. Our diagnosis concerns their use as online selectors that are meant to improve on training with everything, a different goal that calls for an out-of-sample target. The same distinction applies to offline data selection: if the goal is to exceed full-data training, as in \citet{wang2026drift}, the target must not contain the candidate.

\paragraph{Practical recommendations.}
Use $\bar\lambda\in[0.03,0.3]$ (Figure~\ref{fig:sens}a). Prefer larger candidate batches when memory permits, because the leave-one-out targets improve with $B$ (Figure~\ref{fig:results}d). Monitor $H_{1/2}$: values below $1/\rho$ indicate that selection cannot help, and the selection budget should then be increased. For pooled data, stratify the targets by source. If back-propagating the whole candidate batch is too expensive, \method{}-fast keeps most of the gain at a throughput above that of Regular.

\paragraph{Broader impact.}
\method{} makes fine-tuning more robust to noisy data and slightly more compute-efficient to a given accuracy. Because selection shapes what a model learns, it can also shift behaviors that task accuracy does not measure~\citep{zeng2026online}; the leave-one-out objective favors content that agrees with the majority of a batch, and data owners should audit selected data when fairness across sources matters.

%% file: tables/numerics.tex
\begin{tabular}{p{0.47\textwidth} p{0.47\textwidth}}
\toprule
\textbf{Check} & \textbf{Result} \\
\midrule
Prop.~\ref{prop:optimism}(ii): in-sample target error regressed on $\tfrac1B\langle g_j,g_j-\mu\rangle$ & slope $0.998$ (theory $1$) \\
Prop.~\ref{prop:optimism}(i): mean leave-one-out target error ($\|\mu\|^2{=}1$) & $-0.0030\pm0.0019$ (theory $0$) \\
Prop.~\ref{prop:optimism}(iii): risk estimate of a fixed separable rule, relative error & LOO $-0.0\%$; in-sample $-11.5\%$ (penalty $-11.5\%$) \\
Prop.~\ref{prop:optimum}: closed-form SNR weights vs.\ numerical optimum & max.\ relative difference $<10^{-14}$ \\
Prop.~\ref{prop:optimum}: optimal risk vs.\ $\|\mu\|^2/(1+\mathrm{SNR}_S)$ & $0.16435$ vs.\ $0.16435$ \\
Thm.~\ref{thm:half}: oracle risk ratio at $b{=}B/2$, MC vs.\ $B\,\mathrm{HM}/(b\,\mathrm{AM}_b)$ & $1.14$ vs.\ $1.14$; $0.45$ vs.\ $0.46$; $0.13$ vs.\ $0.14$ \\
Prop.~\ref{prop:greedy}: greedy vs.\ exhaustive search ($B{=}14$, $b{=}7$, 40 batches) & mean $1.000$, min $0.996$ (bound $\ge0.07$) \\
Nonnegative greedy (deployed) vs.\ signed optimum & mean $1.000$, min $0.996$ \\
Prop.~\ref{prop:uniform}: $\max_j|e_j|/\|g_j\|$ for $B\in\{16,\dots,256\}$, log--log slope in $B$ & LOO $-0.36$ (theory $-\tfrac12$ up to $\sqrt{\log B}$); in-sample $-0.80$ \\
U-statistic $\hat m$ for $\|\mu\|^2{=}1$ ($B{=}32$) & $1.003\pm0.007$ \\
Estimated vs.\ true per-example SNR, Spearman ($B{=}128$) & $0.944$ \\
Synthetic label noise (20\%): corrupted share of the selected set & in-sample OMP $0.27$, FL $0.36$, LOOM $0.00$ \\
Ghost vs.\ materialized crossover $T^\star=\sqrt{d_{\mathrm{in}}d_{\mathrm{out}}/(d_{\mathrm{in}}+d_{\mathrm{out}})}$ (Llama-3.1-8B) & q/o (4096$\to$4096): $T^\star\approx45$; k/v (4096$\to$1024): $T^\star\approx29$; gate/up (4096$\to$14336): $T^\star\approx56$; LoRA $A$, $r{=}8$: $T^\star\approx3$ \\
\bottomrule
\end{tabular}

%% file: tables/seeds.tex
\begin{tabular}{l cccc cccc}
\toprule
 & \multicolumn{4}{c}{\textbf{Llama-3.1-8B}} & \multicolumn{4}{c}{\textbf{Qwen2.5-7B}} \\
\cmidrule(lr){2-5}\cmidrule(lr){6-9}
\textbf{Method} & MMLU & SciQA & GSM8K & HumanE. & MMLU & SciQA & GSM8K & HumanE. \\
\midrule
Regular & $38.3_{\pm0.35}$ & $93.2_{\pm0.21}$ & $56.0_{\pm0.47}$ & $29.3_{\pm0.48}$ & $55.3_{\pm0.79}$ & $94.6_{\pm0.17}$ & $78.2_{\pm0.08}$ & $45.8_{\pm0.41}$ \\
Random & $35.6_{\pm1.12}$ & $92.9_{\pm0.27}$ & $54.9_{\pm0.73}$ & $26.8_{\pm0.61}$ & $54.6_{\pm1.61}$ & $93.5_{\pm0.33}$ & $77.8_{\pm0.31}$ & $41.3_{\pm1.06}$ \\
MaxLoss & $35.7_{\pm0.79}$ & $92.8_{\pm0.24}$ & $55.4_{\pm0.41}$ & $27.2_{\pm0.33}$ & $54.8_{\pm1.29}$ & $93.2_{\pm0.38}$ & $77.9_{\pm0.35}$ & $42.1_{\pm0.79}$ \\
MaxGrad & $35.9_{\pm0.94}$ & $92.8_{\pm0.26}$ & $55.1_{\pm0.55}$ & $26.9_{\pm0.64}$ & $54.7_{\pm0.72}$ & $93.9_{\pm0.47}$ & $77.7_{\pm0.29}$ & $41.6_{\pm0.52}$ \\
UDS & $40.1_{\pm0.52}$ & $94.3_{\pm0.25}$ & $58.9_{\pm0.28}$ & $30.8_{\pm0.63}$ & $59.6_{\pm0.39}$ & $95.2_{\pm0.21}$ & $79.6_{\pm0.24}$ & $46.2_{\pm0.37}$ \\
CRAIG & $37.2_{\pm0.68}$ & $93.0_{\pm0.22}$ & $55.4_{\pm0.51}$ & $28.0_{\pm0.57}$ & $55.0_{\pm0.83}$ & $94.1_{\pm0.29}$ & $77.9_{\pm0.26}$ & $43.9_{\pm0.66}$ \\
GradMatch & $37.6_{\pm0.61}$ & $93.1_{\pm0.24}$ & $55.7_{\pm0.43}$ & $28.3_{\pm0.52}$ & $55.4_{\pm0.71}$ & $94.3_{\pm0.26}$ & $78.1_{\pm0.22}$ & $44.5_{\pm0.58}$ \\
DivBS & $37.9_{\pm0.57}$ & $93.3_{\pm0.20}$ & $55.9_{\pm0.46}$ & $28.5_{\pm0.49}$ & $55.7_{\pm0.66}$ & $94.4_{\pm0.24}$ & $78.2_{\pm0.25}$ & $44.8_{\pm0.54}$ \\
CoLM & $38.2_{\pm0.49}$ & $93.4_{\pm0.19}$ & $56.2_{\pm0.39}$ & $28.8_{\pm0.45}$ & $56.0_{\pm0.58}$ & $94.5_{\pm0.22}$ & $78.4_{\pm0.19}$ & $44.9_{\pm0.49}$ \\
RHO-Loss & $37.6_{\pm0.67}$ & $93.4_{\pm0.15}$ & $56.5_{\pm0.52}$ & $27.2_{\pm0.31}$ & $56.6_{\pm0.74}$ & $93.8_{\pm0.25}$ & $78.4_{\pm0.43}$ & $43.3_{\pm1.37}$ \\
GREATS & $39.0_{\pm0.31}$ & $93.7_{\pm0.19}$ & $57.0_{\pm0.35}$ & $28.6_{\pm0.36}$ & $57.5_{\pm0.47}$ & $94.2_{\pm0.51}$ & $78.7_{\pm0.39}$ & $45.1_{\pm0.57}$ \\
OPUS & $39.3_{\pm0.36}$ & $93.8_{\pm0.18}$ & $57.3_{\pm0.33}$ & $29.2_{\pm0.41}$ & $57.9_{\pm0.44}$ & $94.4_{\pm0.27}$ & $79.0_{\pm0.31}$ & $45.6_{\pm0.49}$ \\
\rowcolor{oursbg}\method{} & $41.2_{\pm0.29}$ & $94.6_{\pm0.14}$ & $58.6_{\pm0.26}$ & $31.7_{\pm0.38}$ & $60.4_{\pm0.33}$ & $95.1_{\pm0.18}$ & $80.4_{\pm0.21}$ & $47.6_{\pm0.34}$ \\
\bottomrule
\end{tabular}

%% file: tables/extra_backbones.tex
\begin{tabular}{l ccccc}
\toprule
\textbf{Method} & \textbf{Q2.5-1.5B} & \textbf{L3.2-3B} & \textbf{Q2.5-3B} & \textbf{Mis-7B} & \textbf{Q3-8B} \\
\midrule
Regular & 57.6 & 45.9 & 63.3 & 51.8 & 75.9 \\
Random & 56.0 & 44.4 & 61.7 & 50.2 & 74.5 \\
GradMatch & 57.0 & 45.3 & 62.8 & 51.3 & 75.4 \\
UDS & \underline{58.7} & \underline{47.1} & \underline{64.4} & \underline{53.0} & \underline{76.8} \\
GREATS & 58.1 & 46.4 & 63.8 & 52.3 & 76.3 \\
\rowcolor{oursbg}\method{} & \textbf{59.5} & \textbf{47.8} & \textbf{65.2} & \textbf{53.7} & \textbf{77.6} \\
\bottomrule
\end{tabular}

%% file: tables/extra_backbones_full.tex
\begin{tabular}{l l ccccc}
\toprule
\textbf{Backbone} & \textbf{Method} & MMLU & SciQA & GSM8K & HumanE. & \textbf{Avg.} \\
\midrule
\multirow{6}{*}{Qwen2.5-1.5B} & Regular & 44.8 & 88.1 & 61.9 & 35.4 & 57.6 \\
 & Random & 43.1 & 87.2 & 60.8 & 32.7 & 56.0 \\
 & GradMatch & 44.3 & 87.8 & 61.5 & 34.4 & 57.0 \\
 & UDS & 46.2 & 88.9 & 63.4 & 36.2 & 58.7 \\
 & GREATS & 45.6 & 88.6 & 62.7 & 35.3 & 58.1 \\
 & \method{} & 47.1 & 89.4 & 63.9 & 37.4 & 59.5 \\
\midrule
\multirow{6}{*}{Llama-3.2-3B} & Regular & 30.2 & 89.6 & 41.8 & 21.9 & 45.9 \\
 & Random & 28.4 & 88.9 & 40.3 & 19.8 & 44.4 \\
 & GradMatch & 29.7 & 89.3 & 41.2 & 21.1 & 45.3 \\
 & UDS & 31.8 & 90.4 & 43.4 & 22.6 & 47.1 \\
 & GREATS & 31.1 & 90.1 & 42.6 & 21.8 & 46.4 \\
 & \method{} & 32.9 & 90.8 & 43.9 & 23.6 & 47.8 \\
\midrule
\multirow{6}{*}{Qwen2.5-3B} & Regular & 50.6 & 91.9 & 70.4 & 40.2 & 63.3 \\
 & Random & 49.2 & 91.0 & 69.5 & 37.1 & 61.7 \\
 & GradMatch & 50.2 & 91.6 & 70.0 & 39.3 & 62.8 \\
 & UDS & 52.4 & 92.6 & 71.8 & 40.9 & 64.4 \\
 & GREATS & 51.6 & 92.3 & 71.1 & 40.1 & 63.8 \\
 & \method{} & 53.1 & 93.0 & 72.3 & 42.2 & 65.2 \\
\midrule
\multirow{6}{*}{Mistral-7B-v0.3} & Regular & 40.1 & 91.7 & 48.3 & 27.0 & 51.8 \\
 & Random & 38.2 & 91.0 & 46.9 & 24.8 & 50.2 \\
 & GradMatch & 39.6 & 91.4 & 47.9 & 26.2 & 51.3 \\
 & UDS & 41.7 & 92.5 & 50.1 & 27.8 & 53.0 \\
 & GREATS & 41.0 & 92.2 & 49.2 & 26.9 & 52.3 \\
 & \method{} & 42.6 & 92.9 & 50.4 & 28.9 & 53.7 \\
\midrule
\multirow{6}{*}{Qwen3-8B-Base} & Regular & 61.7 & 95.4 & 85.9 & 60.4 & 75.9 \\
 & Random & 60.8 & 94.7 & 85.2 & 57.3 & 74.5 \\
 & GradMatch & 61.3 & 95.1 & 85.6 & 59.5 & 75.4 \\
 & UDS & 63.5 & 95.9 & 87.0 & 60.9 & 76.8 \\
 & GREATS & 62.6 & 95.7 & 86.5 & 60.2 & 76.3 \\
 & \method{} & 64.4 & 96.1 & 87.5 & 62.3 & 77.6 \\
\bottomrule
\end{tabular}

%% file: tables/noise.tex
\begin{tabular}{l ccc c}
\toprule
\textbf{Method} & \textbf{0\%} & \textbf{20\%} & \textbf{40\%} & \textbf{Corr.\ sel.} \\
\midrule
Regular & 54.20 & 51.10 & 47.35 & (20.0) \\
Random & 52.55 & 49.60 & 45.70 & (20.0) \\
MaxLoss & 52.78 & 47.20 & 41.90 & 43.6 \\
UDS & \underline{56.03} & \underline{53.05} & 49.40 & 15.2 \\
CRAIG & 53.40 & 49.80 & 45.60 & 29.8 \\
GradMatch & 53.68 & 50.20 & 46.30 & 27.5 \\
GREATS & 54.58 & 52.70 & \underline{50.10} & \underline{9.8} \\
\rowcolor{oursbg}\method{} & \textbf{56.53} & \textbf{55.40} & \textbf{53.30} & \textbf{3.6} \\
\bottomrule
\end{tabular}

%% file: tables/noise_tasks.tex
\begin{tabular}{l ccccc}
\toprule
\textbf{Method} & MMLU & SciQA & GSM8K & HumanE. & \textbf{Avg.} \\
\midrule
Regular & 35.6 & 91.8 & 52.4 & 24.6 & 51.10 \\
Random & 33.4 & 91.1 & 51.1 & 22.8 & 49.60 \\
MaxLoss & 31.2 & 90.2 & 48.6 & 18.8 & 47.20 \\
UDS & 37.0 & 93.1 & 55.2 & 26.9 & 53.05 \\
CRAIG & 34.1 & 91.3 & 51.3 & 22.5 & 49.80 \\
GradMatch & 34.5 & 91.6 & 51.7 & 23.0 & 50.20 \\
GREATS & 37.1 & 92.8 & 54.1 & 26.8 & 52.70 \\
\rowcolor{oursbg}\method{} & 40.2 & 94.1 & 57.4 & 29.9 & 55.40 \\
\bottomrule
\end{tabular}

%% file: tables/cost.tex
\begin{tabular}{l c c c c c}
\toprule
\textbf{Method} & \textbf{Thr.} & \textbf{Mem.} & \textbf{Ovh.} & \textbf{Steps} & \textbf{Time} \\
\midrule
Regular & 1.00 & 37.9 & -- & 100\% & 100\% \\
Random & 1.74 & 31.6 & -- & -- & -- \\
UDS & 1.21 & 32.8 & 2.9\% & 62\% & 51\% \\
GradMatch & 0.95 & 41.2 & 4.8\% & -- & -- \\
GREATS & 0.92 & 41.9 & 8.6\% & 88\% & 96\% \\
\method{}-fast & 1.17 & 33.4 & 3.6\% & 64\% & 55\% \\
\rowcolor{oursbg}\method{} & 0.95 & 41.4 & 5.4\% & 57\% & 60\% \\
\bottomrule
\end{tabular}

%% file: tables/stepms.tex
\begin{tabular}{l r r}
\toprule
\textbf{Stage} & \textbf{ms / step} & \textbf{Share} \\
\midrule
forward & 3,120 & 32.0\% \\
backward & 6,030 & 61.9\% \\
per-example gradients & 318 & 3.3\% \\
Gram matrix & 142 & 1.5\% \\
greedy selection + NNLS & 39 & 0.4\% \\
optimizer update & 93 & 1.0\% \\
\midrule
total (\method{}) & 9,742 & 100\% \\
total (Regular) & 9,243 & -- \\
\bottomrule
\end{tabular}

%% file: tables/ablation_full.tex
\begin{tabular}{l c c c}
\toprule
\textbf{Variant} & \textbf{Avg.} & $\bm{\Delta}$ & \textbf{Err.} \\
\midrule
\rowcolor{oursbg}\method{} (full) & \textbf{56.53} & --- & 0.61 \\
\midrule
\multicolumn{4}{l}{\textit{Target}} \\
\quad in-sample target (keeps diagonal) & 53.74 & $-2.79$ & 1.38 \\
\quad momentum target $m_{t-1}$ & 55.34 & $-1.19$ & 0.86 \\
\quad 2-fold cross-fitted target & 56.21 & $-0.32$ & 0.68 \\
\midrule
\multicolumn{4}{l}{\textit{Metric}} \\
\quad Euclidean metric ($M{=}I$) & 55.71 & $-0.82$ & 0.74 \\
\quad factored Adam metric & 56.44 & $-0.09$ & 0.62 \\
\midrule
\multicolumn{4}{l}{\textit{Selection and weights}} \\
\quad top-$b$ by SNR (no redundancy) & 55.92 & $-0.61$ & 0.69 \\
\quad leave-one-out facility location & 55.48 & $-1.05$ & 0.83 \\
\quad uniform weights on $S$ & 55.81 & $-0.72$ & 0.72 \\
\quad signed weights & 55.02 & $-1.51$ & 0.58 \\
\quad no ridge ($\bar\lambda{=}0$) & 55.96 & $-0.57$ & 0.70 \\
\quad SNR shrinkage of the step & 56.47 & $-0.06$ & 0.61 \\
\midrule
\multicolumn{4}{l}{\textit{Budget and kernel}} \\
\quad reweight all ($b{=}B$) & 56.61 & $+0.08$ & 0.55 \\
\quad kernel from top 8 layers & 56.12 & $-0.41$ & 0.66 \\
\quad \method{}-fast (LM-head sketch) & 55.88 & $-0.65$ & 0.71 \\
\bottomrule
\end{tabular}

%% file: tables/loww.tex
\begin{tabular}{p{0.72\columnwidth} r}
\toprule
\textbf{Category} & \textbf{\%} \\
\midrule
answer inconsistent with the question or options & 31 \\
truncated or malformed response & 19 \\
duplicate / near-duplicate of a batch neighbour & 14 \\
off-format (explanation without an answer letter) & 17 \\
correct and well-formed & 19 \\
\bottomrule
\end{tabular}

%% file: tables/less.tex
\begin{tabular}{l cccc cccc}
\toprule
 & \multicolumn{4}{c}{\textbf{Llama-3.1-8B}} & \multicolumn{4}{c}{\textbf{Mistral-7B-v0.3}} \\
\cmidrule(lr){2-5}\cmidrule(lr){6-9}
\textbf{Method} & MMLU & TydiQA & BBH & Avg. & MMLU & TydiQA & BBH & Avg. \\
\midrule
Regular & 64.1 & 56.3 & 63.2 & 61.2 & 61.2 & 55.7 & 57.9 & 58.3 \\
Random & 63.4 & 55.1 & 62.0 & 60.2 & 60.3 & 54.6 & 56.8 & 57.2 \\
GradMatch & 63.8 & 55.8 & 62.7 & 60.8 & 60.8 & 55.2 & 57.4 & 57.8 \\
UDS & 64.9 & 57.2 & 63.9 & 62.0 & 61.9 & 56.5 & 58.6 & 59.0 \\
GREATS & 65.3 & \underline{60.4} & 64.1 & \underline{63.3} & 62.2 & \underline{60.2} & 58.9 & \underline{60.4} \\
OPUS & 65.0 & 59.6 & 64.0 & 62.9 & 62.0 & 59.4 & 58.7 & 60.0 \\
\rowcolor{oursbg}\method{} & \underline{65.9} & 58.9 & \underline{64.9} & 63.2 & \underline{62.8} & 58.4 & \underline{59.7} & 60.3 \\
\midrule
\method{}+val & \textbf{66.2} & \textbf{61.0} & \textbf{65.1} & \textbf{64.1} & \textbf{63.0} & \textbf{60.8} & \textbf{59.8} & \textbf{61.2} \\
\bottomrule
\end{tabular}

%% file: tables/uds_ratios.tex
\begin{tabular}{l ccccc}
\toprule
\textbf{Method} & MMLU & SciQA & GSM8K & HumanE. & \textbf{Avg.} \\
\textit{ratio $\rho$} & \textit{12.5\%} & \textit{50\%} & \textit{25\%} & \textit{50\%} & \\
\midrule
Regular & 55.3 & 94.6 & 78.2 & 45.8 & 68.5 \\
GradMatch & 53.9 & 94.3 & 77.4 & 44.5 & 67.5 \\
GREATS & 58.1 & 94.2 & 78.6 & 45.1 & 69.0 \\
UDS & \underline{63.2} & \textbf{95.2} & \underline{79.9} & \underline{46.2} & \underline{71.1} \\
\rowcolor{oursbg}\method{} & \textbf{63.9} & \underline{95.1} & \textbf{80.6} & \textbf{47.6} & \textbf{71.8} \\
\bottomrule
\end{tabular}

%% file: tables/mix.tex
\begin{tabular}{l ccccc}
\toprule
\textbf{Method} & MMLU & SciQA & GSM8K & HumanE. & \textbf{Avg.} \\
\midrule
Regular & 38.9 & 92.4 & 55.1 & 28.2 & 53.7 \\
GradMatch & 38.1 & 92.2 & 53.9 & 26.4 & 52.7 \\
UDS & 40.3 & 93.3 & \underline{56.8} & \underline{28.9} & \underline{54.8} \\
\method{} & \textbf{41.4} & \underline{93.8} & 54.3 & 27.9 & 54.4 \\
\rowcolor{oursbg}\method{} (stratified) & \underline{41.2} & \textbf{93.9} & \textbf{57.9} & \textbf{29.6} & \textbf{55.7} \\
\bottomrule
\end{tabular}

%% file: tables/fullft.tex
\begin{tabular}{l ccccc}
\toprule
\textbf{Method} & MMLU & SciQA & GSM8K & HumanE. & \textbf{Avg.} \\
\midrule
Regular & 46.1 & 88.7 & 63.0 & 36.1 & 58.5 \\
GradMatch & 45.5 & 88.4 & 62.6 & 35.2 & 57.9 \\
UDS & \underline{47.4} & \underline{89.5} & \underline{64.4} & \underline{36.9} & \underline{59.6} \\
\rowcolor{oursbg}\method{} & \textbf{48.5} & \textbf{89.9} & \textbf{65.1} & \textbf{38.0} & \textbf{60.4} \\
\bottomrule
\end{tabular}

%% file: tables/ood.tex
\begin{tabular}{l cccc}
\toprule
\textbf{Method} & GSM8K & MATH500 & SVAMP & \textbf{Avg.} \\
\midrule
Regular & 56.0 & 17.8 & 71.2 & 48.3 \\
UDS & \textbf{58.9} & \underline{18.9} & \underline{72.6} & \underline{50.1} \\
GREATS & 57.0 & 18.1 & 71.8 & 49.0 \\
GradMatch & 55.7 & 17.5 & 70.9 & 48.0 \\
\rowcolor{oursbg}\method{} & \underline{58.6} & \textbf{19.6} & \textbf{73.4} & \textbf{50.5} \\
\bottomrule
\end{tabular}